\documentclass{article} % For LaTeX2e
\usepackage[preprint]{colm2026_conference}

\usepackage{microtype}
\usepackage{hyperref}
\usepackage{url}
\usepackage{booktabs}
\usepackage{graphicx}
\usepackage{xcolor}
\usepackage{colortbl}
\usepackage[table]{xcolor}
\usepackage{lineno} 
\usepackage{amsmath}
\usepackage{tcolorbox}
\usepackage{listings}
\usepackage{xcolor}
\usepackage{enumitem}
\usepackage[T1]{fontenc}
 
\tcbuselibrary{skins, breakable}
 
\newtcolorbox{promptbox}[1]{
  enhanced,
  breakable,
  colback=white,
  colframe=gray!60!black,
  colbacktitle=gray!50,
  coltitle=black,
  fonttitle=\small\bfseries,
  title={#1},
  boxrule=0.8pt,
  arc=2pt,
  left=4pt,
  right=4pt,
  top=4pt,
  bottom=4pt,
  toptitle=2pt,
  bottomtitle=2pt,
  before upper={\setlength{\parskip}{4pt}\setlength{\parindent}{0pt}\sloppy\footnotesize\linespread{1.1}\selectfont},
}
\definecolor{darkblue}{rgb}{0, 0, 0.5}
\definecolor{deltagreen}{rgb}{0.0, 0.45, 0.0}
\definecolor{deltared}{rgb}{0.75, 0.0, 0.0}
\newcommand{\up}[1]{{\scriptsize\textcolor{deltagreen}{\,\textsuperscript{+#1}}}}
\newcommand{\dn}[1]{{\scriptsize\textcolor{deltared}{\,\textsuperscript{#1}}}}
\hypersetup{colorlinks=true, citecolor=darkblue, linkcolor=darkblue, urlcolor=darkblue}

\newcommand{\lightrule}{\arrayrulecolor{black!20}\midrule\arrayrulecolor{black}}

\title{Reflective Context Learning: Studying the Optimization Primitives of Context Space}

\author{Nikita Vassilyev, William Berrios, Ruowang Zhang, Bo Han, Douwe Kiela, Shikib Mehri \\
Contextual AI \\
}

\newcommand{\rcl}{\textsc{RCL}}

\begin{document}

\ifcolmsubmission
\linenumbers
\fi

\maketitle

\renewcommand{\thefootnote}{}
\footnotetext{Code available at \url{https://github.com/nvassilyev/RCL}}
\renewcommand{\thefootnote}{\arabic{footnote}}

\begin{abstract}
Generally capable agents must learn from experience in ways that generalize across tasks and environments. The fundamental problems of learning, including credit assignment, overfitting, forgetting, local optima, and high-variance learning signals, persist whether the learned object lies in parameter space or context space. While these challenges are well understood in classical machine learning optimization, they remain underexplored in context space, leading current methods to be fragmented and ad hoc. We present Reflective Context Learning (RCL), a unified framework for agents that learn through repeated interaction, reflection on behavior and failure modes, and iterative updates to context. In RCL, reflection converts trajectories and current context into a directional update signal analogous to gradients, while mutation applies that signal to improve future behavior in context space. We recast recent context-optimization approaches as instances of this shared learning problem and systematically extend them with classical optimization primitives, including batching, improved credit-assignment signal, auxiliary losses, failure replay, and grouped rollouts for variance reduction. On AppWorld, BrowseComp+, and RewardBench2, these primitives improve over strong baselines, with their relative importance shifting across task regimes. We further analyze robustness to initialization, the effects of batch size, and the impact of allocating stronger or weaker models to different optimization components. Our results suggest that learning through context updates should be treated not as a set of isolated algorithms, but as an optimization problem whose mechanisms can be studied systematically and improved through transferable principles. %Our code has been released at \url{https://github.com/nvassilyev/RCL}
\end{abstract}

\section{Introduction}
\label{sec:intro}

Generally capable agents must learn from experience in ways that generalize across tasks and environments. No fixed model or policy, however capable at training time, can anticipate every environment, user preference, or edge case it will encounter in deployment, and adaptation from online experience is a prerequisite for robust, general-purpose agents~\citep{sutton2018reinforcement, fang2025selfevolving}. The standard approach to adaptation is to update model weights through gradient-based optimization, but this requires expensive retraining, risks catastrophic forgetting, and is difficult to apply continuously during agent operation. An increasingly viable alternative is to optimize the agent's \textit{context} instead: the interpretable artifacts that shape behavior at inference time, such as prompts, structured playbooks, persistent memory, learned tools, and behavioral guidelines \citep{mei2025surveycontextengineering}. Context can be optimized iteratively — through reflection on execution experience rather than backpropagation over a loss surface — without modifying the underlying model. These updates require no gradient computation, can be applied continuously as the agent accumulates experience, and produce human-readable artifacts that can be inspected and debugged. As models become stronger reasoners and more faithful instruction followers, context-space optimization becomes both more effective and more significant: an optimizer that reasons about agent behavior can produce increasingly meaningful, targeted revisions to the agent's operational policy \citep{wang2023voyager, li2025survey}.

% The fundamental difficulties of learning do not disappear when the learned object moves from weights to context. Local optima and overfitting under greedy improvement \citep{mitchell1980biases}, high-variance updates from limited or noisy samples \citep{williams1992simple, greensmith2004variance}, catastrophic forgetting from sequential updates \citep{mccloskey1989catastrophic, kirkpatrick2017overcoming}, and inefficient use of sparse, delayed feedback \citep{sutton2018reinforcement} are properties of iterative optimization under partial information, not of any particular optimization medium. As we demonstrate, these failure modes arise equally when the optimization target is discrete context rather than continuous weights. The classical remedies — batching for variance reduction, replay for forgetting, momentum for stability, structural regularization for generalization \citep{kingma2015adam, schaul2016prioritized, bengio2009curriculum} — are motivated by properties of learning itself, not by properties of any particular optimization substrate, and they transfer to context space when appropriately adapted.

The fundamental difficulties of learning do not disappear when the learned object moves from weights to context. Local optima and overfitting under greedy improvement \citep{mitchell1980biases}, high-variance updates from limited or noisy samples \citep{williams1992simple, greensmith2004variance}, catastrophic forgetting from sequential updates \citep{mccloskey1989catastrophic, kirkpatrick2017overcoming}, and inefficient use of sparse, delayed feedback \citep{sutton2018reinforcement} are properties of iterative optimization under partial information, not of any particular optimization medium. As we demonstrate, these failure modes arise equally in discrete context spaces: a playbook rewritten from one anomalous failure oscillates rather than converges, a fix for a new edge case overwrites a previously mastered strategy, and a reflector that reasons over entire trajectories rather than critical decisions wastes its signal on noise. The classical remedies developed for these problems — batching for variance reduction, replay for forgetting, momentum for stability, and structural regularization for generalization \citep{kingma2015adam, schaul2016prioritized, bengio2009curriculum} — are motivated by properties of learning itself, not by properties of any particular optimization substrate. In parameter-space optimization, the gradient is the mechanism that converts data and a loss signal into a directed update. In context-space learning, this role is played by \textit{reflection}: reasoning over execution trajectories and the current context to diagnose what should change and why. This reflective operation — producing a directional signal from (context, trajectory, outcome) — is what distinguishes learning from search in context space, just as computing gradients distinguishes gradient descent from random search in weight space. When we refer to optimization primitives in this paper, we mean mechanisms that improve the quality, stability, or efficiency of this reflective update process.

% A growing number of methods have begun developing variants of a shared loop — \textit{(i) attempt tasks, (ii) reflect on outcomes, (iii) update context} — introducing mechanisms such as batched textual gradients \citep{pryzant2023automatic}, modular credit assignment \citep{yuksekgonul2024textgrad}, persistent structured memory \citep{zhang2026ace, suzgun2026dynamic}, verbal self-critique \citep{shinn2023reflexion}, grouped rollouts with contrastive advantage \citep{cai2025trainingfreegrpo}, evolutionary search \citep{agrawal2025gepa}, optimizer-state retention \citep{yan2025efficient}, and sampling-based momentum \citep{ding2025samplingmomentum}. Yet these methods were introduced across a period of rapid change in models, prompting practices, and benchmarks, making it difficult to isolate the contribution of a learning primitive from the effects of model capability, prompt engineering, and task difficulty. The problem is compounded by a regime shift: early prompt optimization was largely concerned with conjuring the right tokens to elicit latent capabilities from imperfect instruction followers \citep{pryzant2023automatic, zhou2023ape}, whereas current models are sophisticated enough that context updates can encode genuine strategies learned from ground-truth rewards and execution experience \citep{trivedi2024appworld, wei2025browsecompsimplechallengingbenchmark}. In this regime, the bottleneck moves from search breadth to reflection quality, and which primitives matter, and how they compose, becomes urgent.

A growing number of methods have begun developing variants of a shared loop — attempt tasks, reflect on outcomes, update context — introducing increasingly sophisticated mechanisms along the way. These include batched textual gradients \citep{pryzant2023automatic}, modular credit assignment over computation graphs \citep{yuksekgonul2024textgrad}, persistent structured memory with curation rather than compression \citep{zhang2026ace, suzgun2026dynamic}, verbal self-critique as a cross-episode learning signal \citep{shinn2023reflexion}, grouped rollouts with contrastive advantage estimation \citep{cai2025trainingfreegrpo}, Pareto-aware evolutionary search \citep{agrawal2025gepa}, historical feedback retention as optimizer state \citep{yan2025efficient}, and sampling-based momentum for textual gradient descent \citep{ding2025samplingmomentum}. Yet these methods were introduced over a period of rapid change in the field's underlying infrastructure. The models available to early methods were fundamentally different from those available today, as were prevailing prompting practices, context representations, and evaluation benchmarks. This makes it difficult to isolate the contribution of a learning primitive from the effects of model capability, prompt engineering conventions, and task difficulty. Good ideas may appear to fail because the models of their era couldn't execute them, and implementation choices that were necessary workarounds for weaker models may persist in newer methods without re-examination. The problem is compounded by a regime shift. Early prompt optimization was largely concerned with conjuring the right sequence of tokens to elicit latent capabilities from models that were imperfect instruction followers \citep{pryzant2023automatic, zhou2023ape}. In that setting, the optimization problem was closer to black-box function optimization than to learning: the model already possessed the relevant knowledge, and the challenge was finding wording that reliably surfaced it. The potential for context-space learning is now fundamentally different. Current models are generally capable, sophisticated instruction followers with strong reasoning abilities, which means context updates can do more than optimize phrasing — they can encode genuine strategies, learned from ground-truth labels and reward signals, that transfer across task instances. The challenge shifts from eliciting latent knowledge to learning new knowledge: abstracting multi-step strategies from execution experience on tasks like AppWorld \citep{trivedi2024appworld} and BrowseComp+ \citep{chen2025browsecompplusfairtransparentevaluation}. In this regime, the optimization bottleneck moves from search breadth to reflection quality, and the question of which optimization primitives matter, and how they compose, becomes urgent.

These observations motivate treating context-space adaptation not as a collection of isolated algorithms but as a single optimization problem that can be studied systematically. We formalize this view as \textbf{Reflective Context Learning} (\textbf{RCL}), a framework centered on reflection as the core learning mechanism. RCL constrains the design space to the setting we believe is most practically viable and future-proof: strong reasoning models serving as the optimizer, structured playbook representations as the learned artifact, and iterative, SGD-like updates driven by reflective diagnosis of execution trajectories. 
%Within this setting, the framework defines explicit design axes — how context is parameterized, how the reflective update signal is computed, how that signal is applied, and how optimizer state is managed across iterations — and uses them to study which classical optimization primitives transfer to context space and how they compose. 
Within this setting, we study which classical optimization primitives transfer to context space, how they compose, and how their relative importance shifts across task regimes --- varying how the update signal is computed, how it is applied, and how optimizer state is managed across iterations.
We do not claim to introduce the forward-reflect-update loop itself. ProTeGi \citep{pryzant2023automatic} and TextGrad \citep{yuksekgonul2024textgrad} formalized gradient-like textual feedback, and ACE \citep{zhang2026ace} developed structured playbook artifacts for iterative agent improvement. Our contribution is to identify the shared structure these methods converge on, systematically study how classical optimization primitives compose in context space under controlled conditions, and show that the relative importance of these primitives shifts across task regimes. Across AppWorld \citep{trivedi2024appworld}, BrowseComp+ \citep{chen2025browsecompplusfairtransparentevaluation}, and RewardBench2 \citep{malik2025rewardbench2}, reflection-quality primitives yield the most consistent gains, and training dynamics exhibit qualitative patterns — variance-induced oscillation, momentum-stabilized convergence — analogous to their parameter-space counterparts.

As agents take on increasingly complex tasks and operate in increasingly diverse environments, the ability to learn from experience through context updates will become a core capability. We believe the field will benefit from treating this capability the way classical ML treats parameter-space optimization: not as a collection of isolated methods, but as iterative optimization subject to fundamental pathologies — variance, forgetting, local optima — that can be diagnosed systematically and addressed with transferable primitives. %This paper is a step in that direction.

% =========================================================================
% COAUTHOR NOTES / TO-DO: SECTION 2 (REFLECTIVE CONTEXT LEARNING)
% =========================================================================

\section{Reflective Context Learning}
\label{sec:rcl}

\subsection{Problem Setting and Formulation}
\label{sec:rcl:formulation}

We consider a setting in which an agent $\mathcal{A}$, implemented as a ReAct loop \citep{yao2023react} over a base language model, executes tasks by conditioning on both an input $x$ (a query, environment state, or task specification) and a context artifact $\mathcal{C}$. The context artifact is the collection of all interpretable, externalized components that influence the agent's behavior at inference time: structured playbooks of behavioral rules, persistent memory entries, tool definitions, retrieval indices, and operational guidelines \citep{mei2025surveycontextengineering}. $\mathcal{C}$ is a superset of what is typically called a ``prompt'': it includes any artifact the agent can access during execution, whether injected directly into the context window or accessed dynamically through retrieval or tool use. Modern language models are specifically designed to be effective, faithful instruction followers \citep{ouyang2022instructgpt, anthropic2025claude}, and this sensitivity to $\mathcal{C}$ is what makes context-space optimization practical: updates to $\mathcal{C}$ reliably produce corresponding changes in agent behavior.

Given a dataset of tasks $\mathcal{D} = \{(x_i, y_i^*)\}$ with ground-truth labels or a reward function $\mathcal{R}$ that evaluates trajectory quality, the learning problem is to find the context $\mathcal{C}$ that maximizes expected performance:
\begin{equation}
\label{eq:objective}
\mathcal{C}^* = \arg\max_{\mathcal{C}} \; \mathbb{E}_{(x, y^*) \sim \mathcal{D}}\!\left[\mathcal{R}\!\left(\mathcal{A}(x, \mathcal{C}),\; y^*\right)\right]
\end{equation}
We focus on this supervised, experience-driven optimization setting throughout the paper: given a clear training signal and a collection of execution trajectories, how should the context be updated to improve future behavior?

Several prior methods have instantiated variants of this loop. Reflexion~\citep{shinn2023reflexion} demonstrated that an agent can improve across episodes by appending verbal self-critiques to its context. ProTeGi~\citep{pryzant2023automatic} formalized the update signal as a ``textual gradient,'' using minibatches of failures to produce natural-language critiques applied via beam search. TextGrad~\citep{yuksekgonul2024textgrad} generalized this to compound AI systems, treating entire pipelines as computation graphs with textual feedback propagation. ACE~\citep{zhang2026ace} introduced structured, incremental delta updates to modular playbooks. We abstract over these specific implementations to define a general loop with three stages, each analogous to a stage in gradient-based training:

\begin{table*}[t]
\centering
\small
\caption{Functional correspondence between classical gradient-based learning and reflective context learning, with representative prior work that explored each context-space analogue.}
\label{tab:analogy}
% Requires \usepackage{array} in preamble
\begin{tabular}{@{} m{3.4cm} m{4.8cm} m{4.6cm} @{}}
\toprule
\textbf{Classical Concept} & \textbf{RCL Analogue} & \textbf{Prior Work} \\
\midrule
Parameters $\theta$
  & Context artifact $\mathcal{C}$ (playbook, memory, tools, guidelines)
  & All methods below \\
\lightrule
Forward pass $\hat{y}=f_\theta(x)$
  & Trajectory $\tau = \mathcal{A}(x, \mathcal{C})$
  & ReAct \citep{yao2023react}; Voyager \citep{wang2023voyager} \\
\lightrule
Loss $\mathcal{L}(\hat{y}, y^*)$
  & Outcome signal $r = \mathcal{R}(\tau, y^*)$
  & --- \\
\lightrule
Gradient $\nabla_\theta \mathcal{L}$
  & Reflective diagnostic $\Delta = g(\tau, r, \mathcal{C})$
  & ProTeGi \citep{pryzant2023automatic}; TextGrad \citep{yuksekgonul2024textgrad} \\
\lightrule
Optimizer step
  & Context update $\mathcal{C} \leftarrow f(\mathcal{C}, \Delta)$
  & ACE \citep{zhang2026ace} \\
\lightrule
Minibatch $\mathcal{B} \subset \mathcal{D}$
  & Trajectory batch per reflection step
  & ProTeGi \citep{pryzant2023automatic}; TF-GRPO \citep{cai2025trainingfreegrpo} \\
\lightrule
Momentum / Adam
  & Optimizer history + mutation rationales
  & ERM \citep{yan2025efficient}; \citet{ding2025samplingmomentum} \\
\lightrule
Replay buffer
  & Failure replay over historical hard cases
  & Dyn.\ Cheatsheet \citep{suzgun2026dynamic}; ExpeL \citep{zhao2024expel} \\
\lightrule
Architecture choice
  & Context param.\ (flat vs.\ structured)
  & ACE \citep{zhang2026ace}; DSPy \citep{khattab2024dspy} \\
\lightrule
Regularization
  & Structural constraints (diffs, error$\to$rule maps)
  & ACE \citep{zhang2026ace}; MIPRO \citep{opsahl-ong-etal-2024-optimizing} \\
\bottomrule
\end{tabular}
\end{table*}

% Several prior methods have instantiated variants of this loop~\citep{shinn2023reflexion, pryzant2023automatic, yuksekgonul2024textgrad, zhang2026ace}, reviewed in Section~\ref{sec:rcl:prior}. We abstract over their specific implementations to define a general loop with three stages, each analogous to a stage in gradient-based training:

\paragraph{Forward pass (execution).} Agent $\mathcal{A}$ executes a task, producing a trajectory and outcome:
\begin{equation}
\label{eq:forward}
\tau = \mathcal{A}(x, \mathcal{C}_t), \quad r = \mathcal{R}(\tau, y^*)
\end{equation}
The trajectory $\tau$ is a sequence of actions, observations, and intermediate reasoning steps. The outcome $r$ may be binary, scalar, or a structured execution trace. In classical learning, this corresponds to the forward pass $\hat{y} = f_\theta(x)$ and loss computation $\mathcal{L}(\hat{y}, y^*)$.

\paragraph{Backward pass (reflection).} The reflector $g$ takes the trajectory, its outcome, and the current context as input and produces a structured diagnostic signal:
\begin{equation}
\label{eq:backward}
\Delta = g\!\left(\tau, r,\; \mathcal{C}_t\right)
\end{equation}
The diagnostic $\Delta$ is a natural-language analysis of what failed, why, and what components of $\mathcal{C}_t$ should be revised. This is the context-space analogue of gradient computation $\nabla_\theta \mathcal{L}$: converting execution experience into a directed update signal. The key difference is that $\Delta$ is produced by LLM reasoning over trajectories rather than by differentiating through a computation graph; the reflector may be a separate, potentially stronger model.

\paragraph{Optimizer step (mutation).} The mutator $f$ takes the current context and the diagnostic signal and produces an updated context:
\begin{equation}
\label{eq:optim}
\mathcal{C}_{t+1} = f(\mathcal{C}_t, \Delta)
\end{equation}
This is the context-space analogue of the optimizer step $\theta_{t+1} = \theta_t - \alpha \nabla \mathcal{L}$. The mutator applies the reflector's recommendations, subject to structural constraints such as only modifying specific playbook rules or maintaining version history.

\noindent The full update, iterating over samples drawn from $\mathcal{D}$, is:
\begin{equation}
\label{eq:rcl}
\mathcal{C}_{t+1} = f\!\left(\mathcal{C}_t, \; g\!\left(\tau_t, r_t, \mathcal{C}_t\right)\right)
\end{equation}

The analogy to gradient-based optimization is functional, not formal. There is no differentiable loss surface, and $\Delta$ can be wrong, vague, or contradictory in ways that real gradients cannot. We do not claim mathematical equivalence. %What we claim is that the functional roles are preserved (Table~\ref{tab:analogy}), and that the classical mechanisms that improve gradient-based learning improve reflective context learning for the same underlying reasons.
% The analogy is functional, not formal: there is no differentiable loss surface, and $\Delta$ can be wrong or vague in ways real gradients cannot. We claim not mathematical equivalence but that the functional roles are preserved (Table~\ref{tab:analogy}) and that the classical mechanisms that improve gradient-based learning improve reflective context learning for the same underlying reasons.
What we claim is that the functional roles are preserved, as summarized in Table~\ref{tab:analogy}: the reflector serves the same purpose as gradient computation (converting experience into a directed update signal), the mutator serves the same purpose as an optimizer step (applying that signal), and the classical mechanisms that improve gradient-based learning improve reflective context learning for the same underlying reasons. 
When we refer to \textit{optimization primitives} throughout this paper, we mean mechanisms that improve the quality, stability, or efficiency of this reflective update process, which we study in Section~\ref{sec:primitives}.
The independence of $g$, $f$, and $\mathcal{A}$ and the structured nature of $\mathcal{C}$ are deliberate: independent components allow model capacity to be allocated where it matters most, and structured context enables the localized credit assignment and targeted edits that the primitives rely on.

% ============================================================
% 2.2: Merged narrative — origins through current landscape,
%      with directional clustering of improvements.
% 2.3: Design axes (unchanged, tight).
% Old 2.3 (Other Paradigms) folded entirely into 2.2.
% Target: ~1.2 pages total.
% ============================================================

\subsection{Prior Work and the Emergence of Context-Space Learning}
\label{sec:rcl:prior}

Context-space optimization has its roots in prompt engineering and discrete prompt tuning. Soft prompt tuning~\citep{lester2021power, li2021prefix} offered a gradient-based alternative over continuous embeddings, but operates in a space where gradient steps produce micro-level perturbations rather than structured, interpretable revisions. In-context learning~\citep{brown2020language} demonstrated that context is a powerful conditioning mechanism but involves no iterative refinement. Early discrete methods like APE~\citep{zhou2023ape} introduced iteration through generate-and-score search, establishing context as an optimization target, but the update signal remained purely scalar with no diagnosis of why one candidate outperformed another.

The transition to genuine learning began with the introduction of \textit{reflection} as the update mechanism. Reflexion~\citep{shinn2023reflexion} demonstrated that verbal self-critique, appended to context across episodes, improves agent performance without weight updates. ProTeGi~\citep{pryzant2023automatic} formalized this as ``textual gradients'': batched natural-language critiques derived from minibatches of failures and applied via beam search. TextGrad~\citep{yuksekgonul2024textgrad} generalized the gradient metaphor to compound AI systems, treating multi-component pipelines as computation graphs with textual feedback propagation. Together, these methods established that context-space learning shares the structure of gradient-based optimization --- reinforced by concurrent work framing the same loop as reinforcement learning without weight updates~\citep{cai2025trainingfreegrpo,agrawal2025gepa}.

% These methods made explicit that context-space learning shares the structure of gradient-based optimization, with the reflector playing the role of gradient computation and the mutator playing the role of the optimizer step. This connection has been reinforced by concurrent methods that frame context-space adaptation directly as reinforcement learning without weight updates~\citep{cai2025trainingfreegrpo,shao2024deepseekmath, agrawal2025gepa}, further supporting the view that optimization primitives from parameter space transfer to context space.

From this shared foundation, prior work has explored several directions of improvement. \textbf{Structured parameterization:} moving from flat prompts to modular representations enables localized credit assignment; Dynamic Cheatsheet~\citep{suzgun2026dynamic} introduced persistent curated memory, ACE~\citep{zhang2026ace} developed structured playbooks with delta edits, and DSPy~\citep{khattab2024dspy} and MIPRO~\citep{opsahl-ong-etal-2024-optimizing} optimize over modular LM programs. \textbf{Variance reduction:} ProTeGi's minibatching aggregates critiques across failures, and Training-Free GRPO~\citep{cai2025trainingfreegrpo} further reduces noise through grouped rollouts with contrastive semantic advantages. \textbf{Optimizer state and momentum:} ERM~\citep{yan2025efficient} retains historical feedback to prevent information loss,~\citet{ding2025samplingmomentum} introduced sampling-based momentum, and OPRO~\citep{yang2023llmoptimizers} passes optimization history into context. \textbf{Search and frontier maintenance:} EvoPrompt~\citep{guo2024evoprompt} and PromptBreeder~\citep{fernando2024promptbreeder} maintain evolved candidate populations, and GEPA~\citep{agrawal2025gepa} combines population search with reflective diagnosis. \textbf{Policy-level learning:} ExpeL~\citep{zhao2024expel} extracts reusable insights for cross-task transfer, and Agent-Pro~\citep{zhang2024agentpro} revises behavioral beliefs and guidelines.

%todo: uncomment
These directions address three fundamental dimensions that any learning system must navigate: \textit{parameterization} (the structure of the learned artifact determines what granularity of credit assignment is possible), \textit{signal quality} (the precision of the reflective diagnostic depends on how many trajectories the reflector observes and whether it can isolate critical decisions), and \textit{optimizer dynamics} (without momentum, replay, or curriculum, a stateless learner may oscillate, forget, or overfit to recent experience). However, these directions have been explored in isolation, each under particular models, prompting conventions, benchmarks, and task regimes. As discussed in Section~\ref{sec:intro}, the rapid co-evolution of model capabilities and evaluation practices makes it difficult to attribute improvements to specific learning primitives rather than to stronger base models or better prompt engineering. A comprehensive study under controlled conditions --- fixing the base model, context representation, and evaluation protocol while varying the optimization primitives --- is needed to understand which mechanisms matter, how they compose, and how their relative importance shifts across task regimes. Section~\ref{sec:primitives} introduces this study.

%=========================================================================
% COAUTHOR NOTES / TO-DO: SECTION 3 (OPTIMIZATION PRIMITIVES)
% Enforce the 3-part template: 
% 1. Classical Pathology
% 2. Prior Context-Space Work 
% 3. RCL Implementation
% =========================================================================

\section{Optimization Primitives}
\label{sec:primitives}

% Section~\ref{sec:rcl} defined three stages --- execution (Eq.~\ref{eq:forward}), reflection (Eq.~\ref{eq:backward}), and mutation (Eq.~\ref{eq:optim}) --- and argued that classical optimization pathologies should arise in each. We now make this concrete. For each of six primitives (Table~\ref{tab:primitives}), we identify which stage it targets, what pathology it addresses, and how we instantiate it within the RCL loop. Section~\ref{sec:results_analysis} evaluates each primitive individually against the ACE baseline and studies their composition.

\begin{figure*}[t]
\centering
\includegraphics[width=\textwidth]{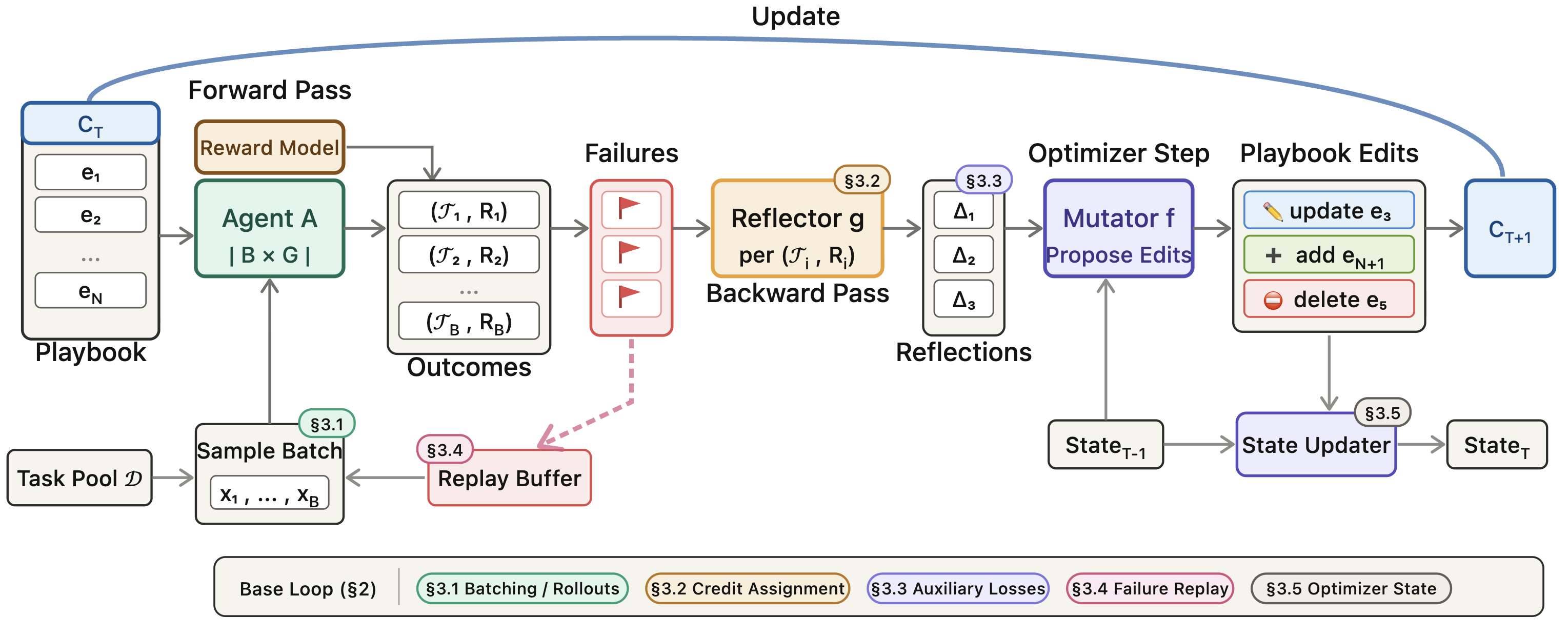}
%\vspace{-2em} %hack for space
\caption{The RCL optimization loop with primitives mapped to stages. A batch of $B$ tasks is sampled from $\mathcal{D}$ (with replay, \S\ref{sec:replay}), each executed $G$ times (\S\ref{sec:batching}). Failed traces are passed --- alongside dual-trace annotations (\S\ref{sec:credit}) --- to a multi-head reflector (\S\ref{sec:aux}) that produces per-trace diagnostics $\Delta_i$. The mutator $f$ aggregates these into structured edits to individual playbook entries $e_i$, conditioned on a rolling optimization state document (\S\ref{sec:momentum}).}
\label{fig:rcl-overview}
\vspace{-1em} %hack for space
\end{figure*}

The core update $\mathcal{C}_{t+1} = f(\mathcal{C}_t, g(\tau_t, r_t, \mathcal{C}_t))$ is a single-sample, stateless, greedy step: one trajectory informs one reflection, which produces one edit, with no memory of prior iterations and no mechanism to escape a poor basin. Applied repeatedly, this minimal loop exhibits the same pathologies as its parameter-space counterpart --- high-variance updates, sparse credit assignment, catastrophic forgetting, and local optima. We introduce five primitives that address these pathologies at specific stages of the loop (Table~\ref{tab:primitives}; Figure~\ref{fig:rcl-overview}). Section~\ref{sec:main_results} evaluates each primitive against the ACE baseline and studies their composition. Each primitive modifies the base update of Eq.~\ref{eq:rcl} in a specific, localizable way. We describe each by stating which component of the update it changes, what pathology motivates it, and how it is instantiated.

\subsection{Batching and Grouped Rollouts}
\label{sec:batching}

A single trajectory $\tau_t$ produces a single diagnostic $\Delta_t$ whose content is dominated by the idiosyncrasies of that example --- the context-space analogue of the high-variance updates that minibatching addresses in parameter-space optimization~\citep{robbins1951stochastic, pryzant2023automatic, cai2025trainingfreegrpo}. We implement two complementary axes of variance reduction.

\paragraph{Task batching.} Instead of reflecting on a single trajectory, we sample $B$ tasks per iteration $\{x_1, \ldots, x_B\} \sim \mathcal{D}$, execute each, and reflect on each failed trace independently, producing per-trace diagnostics $\Delta_1, \ldots, \Delta_k$ for $k \leq B$ failures. These are passed jointly to the mutator:
\begin{equation}
\label{eq:batching}
\mathcal{C}_{t+1} = f\!\left(\mathcal{C}_t, \; \Delta_1, \ldots, \Delta_k\right)
\end{equation}
The mutator identifies recurring patterns across diagnostics and filters one-off anomalies, reducing variance across the task distribution. When $k = 0$ (all tasks pass), no diagnostics are produced and the mutator makes no edit; this is a realistic scenario, particularly on AppWorld where seed scores exceed 78\%. This parallels minibatching in SGD, where averaging gradients over $B$ samples reduces update variance by $\mathcal{O}(1/B)$; here the ``averaging'' is performed by the mutator's reasoning rather than by arithmetic mean.

\paragraph{Grouped rollouts.} Each task $x_i$ is executed $G$ times under the same playbook $\mathcal{C}_t$, yielding a group $\{\tau_i^{(1)}, \ldots, \tau_i^{(G)}\}$ with outcomes $\{r_i^{(1)}, \ldots, r_i^{(G)}\}$. Groups containing both passes and failures provide contrastive signal --- the reflector receives both a successful trace $\tau^+$ and a failed trace $\tau^-$ for the same task:
\begin{equation}
\label{eq:grouped}
\Delta_i = g\!\left(\tau_i^+, \tau_i^-, r_i^+, r_i^-, \; \mathcal{C}_t\right)
\end{equation}
enabling it to isolate the decision points responsible for the outcome difference while controlling for task difficulty. When a group contains no contrastive signal (all $G$ traces pass or all fail), the reflector falls back to the single-trace signature of Eq.~\ref{eq:backward}; the contrastive form of Eq.~\ref{eq:grouped} applies only when both outcomes are present. This is the only setting in which the reflector receives positive traces, grounding its diagnoses in demonstrated successful behavior. Batching reduces variance across the task distribution; grouped rollouts reduce variance within each task, analogous to the distinction between inter-sample and intra-sample variance reduction in Monte Carlo methods.

\subsection{Improved Credit Assignment}
\label{sec:credit}

The outcome signal $r = \mathcal{R}(\tau, y^*)$ is terminal, leaving the reflector to attribute it across an entire trajectory and playbook with no intermediate supervision --- the same sparse-reward problem that motivates step-level reward modeling~\citep{lightman2024lets} and value decomposition~\citep{sunehag2018valuedecomposition}. We address this with \textbf{dual-trace} credit assignment. Let $\mathcal{C}^{\text{ann}}_t$ denote an annotated variant of $\mathcal{C}_t$ that injects XML instrumentation into each entry $e_i$, prompting the agent to cite which entries it consulted, flag uncertainty, and note where guidance was missing. Each task is executed twice concurrently:
\begin{align}
\label{eq:dual-trace}
\tau^{\text{std}} &= \mathcal{A}(x, \mathcal{C}_t), \quad r^{\text{std}} = \mathcal{R}(\tau^{\text{std}}, y^*) \\
\tau^{\text{ann}} &= \mathcal{A}(x, \mathcal{C}^{\text{ann}}_t), \quad r^{\text{ann}} = \mathcal{R}(\tau^{\text{ann}}, y^*) \nonumber
\end{align}
The standard trace $\tau^{\text{std}}$ remains uncontaminated for evaluation; the annotated trace $\tau^{\text{ann}}$ makes the agent's decision process observable, enabling entry-level attribution. The reflector receives both traces but only the standard outcome:
\begin{equation}
\Delta = g\!\left(\tau^{\text{std}}, \tau^{\text{ann}}, r^{\text{std}}, \; \mathcal{C}_t\right)
\end{equation}
The annotated outcome $r^{\text{ann}}$ is excluded because instrumentation alters the agent's behavior, making $r^{\text{ann}}$ an unreliable measure of the playbook's quality; the annotated trace is used only for its decision-process observability, not its outcome. When composed with grouped rollouts, total executions per task become $G + 1$: $G$ baseline traces plus one annotated trace, excluded from the contrastive group because instrumentation alters behavior.
\begin{table}[t]
\centering
\small
\caption{Optimization primitives, the classical pathologies they address, and the RCL loop stage each targets.}
\label{tab:primitives}
\begin{tabular}{@{} p{3.1cm} p{1.4cm} p{4.8cm} p{3.3cm} @{}}
\toprule
\textbf{Primitive} & \textbf{Stage} & \textbf{Pathology} & \textbf{Prior Work} \\
\midrule
Batching & Execution & High variance from single samples & ProTeGi; TF-GRPO \\
\lightrule
Grouped Rollouts & Execution & Confounded attribution & TF-GRPO \\
\lightrule
Credit Assignment & Reflection & Sparse terminal reward & TextGrad \\
\lightrule
Auxiliary Losses & Reflection & Surface-level diagnostics & --- \\
\lightrule
Failure Replay & Sampling & Forgetting learned tactics & Dyn.\ Cheatsheet; ExpeL \\
\lightrule
Optimizer State & Mutation & Oscillation from stateless updates & ERM; OPRO \\
\bottomrule
\end{tabular}
\end{table}

\subsection{Auxiliary Losses and Structural Inductive Biases}
\label{sec:aux}

Without explicit structure, unconstrained reflections collapse toward surface-level trajectory retelling --- analogous to the representation collapse that auxiliary objectives prevent in multi-task learning~\citep{jaderberg2016reinforcement}. We impose structure at two levels.

\paragraph{Playbook parameterization.} $\mathcal{C}$ is organized into named sections with individually addressable entries $e_1, \ldots, e_N$, and the mutator is constrained to express updates as localized edit operations --- $\textsc{Update}(e_j, e_j')$, $\textsc{Add}(e_{N+1})$, $\textsc{Delete}(e_k)$ --- rather than holistic rewrites. This structural constraint is analogous to sparsity regularization in parameter space: it limits the degrees of freedom per update, preventing the mutator from making sweeping changes that overfit to the current batch.

\paragraph{Reflection schema.} The reflector $g$ is decomposed into three parallel diagnostic heads:
\begin{equation}
\label{eq:multihead}
\Delta = \left(\Delta^{\text{attr}}, \; \Delta^{\text{root}}, \; \Delta^{\text{gap}}\right)
\end{equation}
where $\Delta^{\text{attr}} \in \{\textit{actionable gap}, \textit{execution variance}, \textit{intractable}\}$ is a failure attribution classification, $\Delta^{\text{root}}$ is a root cause analysis, and $\Delta^{\text{gap}}$ specifies a coverage gap in the current playbook. These heads interact with the mutator: execution-variance attributions bias toward no-ops (preventing unnecessary edits from noisy signal), while actionable-gap attributions with specific root causes drive targeted additions or modifications. The decomposition forces the reflector to produce structured diagnoses rather than unstructured narrative, analogous to how auxiliary loss heads in multi-task learning force intermediate representations to capture specific aspects of the input.
\subsection{Failure Replay and Curriculum Strategy}
\label{sec:replay}

A single reflection-mutation cycle may not resolve a failure: the edit may be partial, may interact negatively with existing entries $e_i$, or may require refinement only apparent on re-encounter in a different batch context. Experience replay buffers address analogous issues in parameter-space learning~\citep{lin1992selfimproving, schaul2016prioritized}; in context space, where edits can directly contradict or subsume one another, the need is at least as acute.

We maintain a failure replay buffer $\mathcal{B}_t$ that modifies the sampling distribution at each iteration. Let $\rho \in [0, 1]$ be the replay ratio. At each iteration, $\lfloor \rho \times B \rfloor$ tasks are drawn from $\mathcal{B}_t$ and the remaining $B - \lfloor \rho \times B \rfloor$ are sampled fresh from $\mathcal{D}$:
\begin{equation}
\label{eq:replay}
\{x_1, \ldots, x_B\} \sim (1 - \rho) \cdot \mathrm{Uniform}(\mathcal{D}) \;+\; \rho \cdot \mathrm{Uniform}(\mathcal{B}_t)
\end{equation}
Tasks enter $\mathcal{B}_t$ upon failure and are managed by two thresholds: a task \emph{graduates} (is removed) after $n_{\text{grad}}$ consecutive passes across iterations, confirming the playbook has durably learned to handle it; a task is \emph{evicted} after $n_{\text{evict}}$ consecutive failures, indicating it may be intractable under the current playbook and should not dominate the training signal. This implements a curriculum that concentrates optimization effort where the marginal return is highest, analogous to prioritized experience replay~\citep{schaul2016prioritized} where samples are weighted by their learning utility rather than drawn uniformly.

\subsection{Optimizer State and Momentum}
\label{sec:momentum}

A stateless optimizer may revert a change from two iterations ago because the evidence that motivated it has scrolled out of context --- the analogue of the oscillation that momentum~\citep{polyak1964some, kingma2015adam} was designed to prevent. In gradient-based optimization, momentum maintains an exponential moving average $m_t = \beta m_{t-1} + (1-\beta) \nabla \mathcal{L}_t$, smoothing the update trajectory. We implement an analogous mechanism in context space.

We maintain a structured, rolling optimization state document $\mathrm{S}_t$, updated by a dedicated model call $h$ after each iteration:
\begin{equation}
\label{eq:state-update}
\mathrm{S}_{t+1} = h\!\left(\mathrm{S}_t, \; \Delta_1, \ldots, \Delta_k, \; \mathcal{C}_t, \mathcal{C}_{t+1}\right)
\end{equation}
$\mathrm{S}_t$ tracks a change ledger (what was modified and why), playbook assessment (which entries are working well vs.\ poorly), open hypotheses (conjectured failure modes not yet confirmed), and optimization phase (exploratory vs.\ convergent). The state document is injected into the mutator but excluded from the reflector:
\begin{equation}
\label{eq:state-inject}
\mathcal{C}_{t+1} = f\!\left(\mathcal{C}_t, \; \Delta_1, \ldots, \Delta_k, \; \mathrm{S}_t\right)
\end{equation}
This asymmetry mirrors how momentum in Adam operates on the optimizer step rather than on gradient computation: the reflector's diagnostics $\Delta_i$ remain unbiased by the consensus of past iterations, while the mutator can use $\mathrm{S}_t$ to contextualize current diagnostics, avoid reverting previously validated changes, and maintain consistency across iterations.

\paragraph{Full composed update.} Incorporating all five primitives, the RCL update becomes:
\begin{equation}
\label{eq:rcl-full}
\mathcal{C}_{t+1} = f\!\left(\mathcal{C}_t, \; \left\{g\!\left(\tau_i^+, \tau_i^-, \tau_i^{\text{ann}}, r_i, \mathcal{C}_t\right)\right\}_{i \in \mathcal{B}_t^{\rho}}, \; \mathrm{S}_t\right)
\end{equation}
where $\mathcal{B}_t^{\rho}$ denotes the replay-mixed batch of $B$ tasks (Eq.~\ref{eq:replay}), each executed $G+1$ times (grouped rollouts plus one annotated trace), $g$ is the multi-head reflector of Eq.~\ref{eq:multihead}, and $\mathrm{S}_t$ is the optimizer state. Each primitive addresses a distinct pathology at a specific stage of this update; Section~\ref{sec:main_results} evaluates their individual and composed contributions.

\section{Experiments}
\label{sec:experiments}

\subsection{Setup}
\label{sec:setup}

We evaluate on three benchmarks that place different demands on the
optimization loop. \textbf{AppWorld}~\citep{trivedi2024appworld} is a
multi-step interactive coding benchmark scored by Task Goal Completion
(TGC); we sample training trajectories from a pool of 90 tasks and
evaluate on held-out Normal (168 tasks) and Challenge (417 tasks) test
splits. With seed scores of 78--82\% on Normal, the optimization problem
resembles finetuning, where the base model already possesses core
capabilities and gains come from correcting procedural failure modes.
\textbf{BrowseComp+}~\citep{chen2025browsecompplusfairtransparentevaluation}
is a web research benchmark scored by LLM-judged accuracy; we sample from
a training pool of 100 queries, use 30 queries for validation, and
evaluate on 150 held-out queries. Seed scores of 29--41\% mean the problem
is closer to skill acquisition, requiring the model to discover general
search heuristics it does not yet possess.
\textbf{RewardBench2}~\citep{malik2025rewardbench2} is a response-ranking
task scored by accuracy; we sample from a training pool of 1{,}307
examples, validate on 277, and evaluate on a 281-example test split. Seed
scores of 68--76\% in a near-deterministic, non-agentic environment make
this a calibration problem---refining discriminative criteria rather than
learning new procedures or strategies. In all cases, the optimizer sees at
most $B \times T$ training examples over $T$ iterations, where $B$ is the
batch size; coverage of the full training pool is neither required nor
guaranteed, and sampling is governed by the replay buffer when active
(\S\ref{sec:replay}). Rollout budgets reflect the differing seed failure
rates: AppWorld's high seed pass rate requires fewer executions per
iteration to surface failures, while BrowseComp+'s low pass rate demands
larger rollout batches.
% We evaluate on three benchmarks with different optimization demands. \textbf{AppWorld}~\citep{trivedi2024appworld} is a multi-step interactive coding benchmark scored by Task Goal Completion (TGC) on Normal (168 tasks) and Challenge (417 tasks) splits; seed scores of 78--82\% on Normal mean gains come from correcting procedural failure modes. \textbf{BrowseComp+}~\citep{chen2025browsecompplusfairtransparentevaluation} is a web research benchmark scored by LLM-judged accuracy on 150 held-out queries; seed scores of 29--41\% require discovering general search heuristics the model does not yet possess. \textbf{RewardBench2}~\citep{malik2025rewardbench2} is a response-ranking task scored by accuracy on a 280-example test subset; seed scores of 68--76\% in a near-deterministic environment make this a calibration problem.

We use two agent models: Gemini~3.1~Flash~Lite (Lite) and GPT-5.4~Nano (Nano). All runs use Claude Opus 4.6 as reflector and mutator, train for 30 iterations, and are evaluated on held-out splits never seen during training. We compare against two baselines.

\textbf{ACE}~\citep{zhang2026ace} is our primary baseline: it serves as the base optimization loop upon which all of our experiments are built, and corresponds to Eq.~\ref{eq:rcl} with none of the primitives from Section~\ref{sec:primitives} active. We adopt its structured delta edits, helpful/harmful bullet scoring, and the Generator--Reflector--Curator decomposition. We omit two optional mechanisms: (i) the embedding-based de-duplication step, replacing it with explicit update/delete operations that manage playbook size without an external embedding model, and (ii) multi-round Reflector refinement, using a single reflection pass instead. Neither omission changes the fundamental loop; both simplify the infrastructure while keeping the baseline competitive.

\textbf{GEPA}~\citep{agrawal2025gepa} is a sample-efficient prompt optimizer that collects execution traces and applies natural-language reflection to diagnose errors and propose prompt revisions. A genetic Pareto search over candidate prompts maintains a diversity frontier, helping avoid local optima. We use the official DSPy implementation~\citep{khattab2024dspy} with \texttt{auto="light"}, which most closely matches our experimental budget (typically exceeding 30 optimizer iterations), and a mini-batch size of 3, equal to our batch size $B{=}3$. GEPA performs reflective updates on the training set but scores candidates on a held-out validation set to maintain its Pareto frontier; we use 56 tasks for AppWorld, 30 queries for BrowseComp+, and 277 examples for RewardBench2.

Beyond baselines, all methods share the same agent model, optimizer model, and evaluation protocol; only the optimization primitives vary. Each benchmark uses its own seed playbook (Appendix~\ref{app:seed-playbooks}) and system prompt (Appendix~\ref{app:prompts}). The ACE baseline reflects on a single failed trace per iteration ($B{=}1$). The Batching primitive increases this to $B{=}3$; the composed \rcl{} configuration also uses $B{=}3$, with $G{=}3$ grouped rollouts and a replay ratio of $\rho{=}0.5$ across all benchmarks. Section~\ref{sec:seed} studies sensitivity to seed quality.

\paragraph{Rollout protocol.} At each iteration of the ACE loop, the optimizer samples and executes a set of tasks, then selects up to $B$ failed traces for reflection and mutation. The rollout budget is calibrated to the seed failure rate of the current playbook so that each iteration reliably surfaces enough negatives to fill a batch. This calibration matters because long-horizon agentic tasks are expensive to execute: each rollout may involve multi-step tool use, API calls, and environment interaction, making it impractical to sample until failures appear by chance. When grouped rollouts are active, groups are prioritized for selection based on contrastive signal: groups containing at least one pass and one fail are selected first, as these provide the sharpest attribution; if all groups fail uniformly and the evaluation signal is non-binary, we select groups with the largest variance.

\subsection{Main Results}
\label{sec:main_results}

\begin{table*}[!t]
\centering
\small
% \caption{\textbf{Main results.}s across benchmarks and agent models. Each \rcl{} primitive is added individually to the ACE baseline. Best result per column is \textbf{bolded}. Metrics: AppWorld uses Task Goal Completion (TGC); BrowseComp+ and RewardBench2 use accuracy. Agent models: Lite = Gemini~3.1~Flash-Lite, Nano = GPT 5.4-Nano. All runs use Claude Opus as reflector/mutator with 30 training iterations. Deltas are relative to Seed.}
\caption{\textbf{Main results.} Each primitive is added individually to the ACE baseline. AppWorld: TGC; BrowseComp+/RewardBench2: accuracy. Lite = Gemini~3.1~Flash~Lite, Nano = GPT-5.4~Nano. All runs use Claude Opus; 30 training iterations. Deltas vs.\ Seed.}

\label{tab:main-results}
\resizebox{\textwidth}{!}{%
\begin{tabular}{@{} l cc cc cc cc @{}}
\toprule
& \multicolumn{4}{c}{\textbf{AppWorld}} & \multicolumn{2}{c}{\textbf{BrowseComp+}} & \multicolumn{2}{c}{\textbf{RewardBench2}} \\
\cmidrule(lr){2-5} \cmidrule(lr){6-7} \cmidrule(lr){8-9}
& \multicolumn{2}{c}{Normal} & \multicolumn{2}{c}{Challenge} & \multicolumn{2}{c}{Test Set} & \multicolumn{2}{c}{Test Subset} \\
\cmidrule(lr){2-3} \cmidrule(lr){4-5} \cmidrule(lr){6-7} \cmidrule(lr){8-9}
& Lite & Nano & Lite & Nano & Lite & Nano & Lite & Nano \\
\midrule
\multicolumn{9}{@{}l}{\textit{Baselines}} \\[2pt]
Seed & 78.0\phantom{\up{00.0}} & 81.5\phantom{\up{00.0}} & 64.3\phantom{\up{00.0}} & 75.8\phantom{\up{00.0}} & 28.9\phantom{\up{00.0}} & 40.7\phantom{\up{00.0}} & 75.9\phantom{\up{00.0}} & 67.2\phantom{\up{00.0}} \\
GEPA & 82.7\up{4.7} & 76.2\dn{-5.3} & 62.6\dn{-1.7} & 66.7\dn{-9.1} & 35.3\up{6.4} & 48.7{\up{8.0}} & 79.2\up{3.3} & 76.5\up{9.3} \\
ACE & 83.3\up{5.3} & 86.9\up{5.4} & 69.1\up{4.8} & 80.7\up{4.9} & 37.3\up{8.4} & 50.0\up{9.3} & 79.2\up{3.3} & 62.7\dn{-4.5} \\
\midrule
\multicolumn{9}{@{}l}{ACE + \textit{individual primitive}} \\[2pt]
+ Failure Replay & 84.5\up{6.5} & 82.7\up{1.2} & 73.1\up{8.8} & 81.8\up{6.0} & \textbf{40.0}\up{11.1} & 47.3\up{6.6} & 80.4\up{4.5} & 65.4\dn{-1.8} \\
+ Optimizer State & 88.1\up{10.1} & 84.5\up{3.0} & 73.9\up{9.6} & 79.4\up{3.6} & 36.0\up{7.1} & \textbf{56.4}\up{15.7} & 82.8\up{6.9} & 69.9\up{2.7} \\
+ Credit Assignment & 83.9\up{5.9} & 84.5\up{3.0} & 71.5\up{7.2} & 79.4\up{3.6} & 37.3\up{8.4} & 46.3\up{5.6} & 77.2\up{1.3} & 63.5\dn{-3.7} \\
+ Grouped Rollouts & 86.3\up{8.3} & 86.8\up{5.3} & 72.4\up{8.1} & 81.4\up{5.6} & 38.7\up{9.8} & 50.7\up{10.0} & 81.5\up{5.6} & \textbf{77.8}\up{10.6} \\
+ Batching & 88.7\up{10.7} & 87.2\up{5.7} & 71.2\up{6.9} & \textbf{83.4}\up{7.6} & 31.3\up{2.4} & 48.3\up{7.6} & \textbf{82.9}\up{7.0} & 63.6\dn{-3.6} \\
+ Auxiliary Losses & 86.3\up{8.3} & \textbf{89.2}\up{7.7} & \textbf{74.6}\up{10.3} & 80.6\up{4.8} & 38.0\up{9.1} & 54.7\up{14.0} & 82.5\up{6.6} & 71.2\up{4.0} \\
\midrule
\multicolumn{9}{@{}l}{\textit{Composed}} \\[2pt]
\rcl{} (all primitives) & \textbf{89.3}\up{11.3} & 89.1\up{7.6} & 71.9\up{7.6} & \textbf{83.7}\up{7.9} & 37.3\up{8.4} & 51.3\up{10.6} & 82.4\up{6.5} & 69.8\up{2.6} \\
\bottomrule
\end{tabular}%
}
\end{table*}

%Table~\ref{tab:main-results} reports results for each primitive added to ACE, and for the full \rcl{} composition.

\paragraph{Reflection quality gives the best returns per compute.} Optimizer state and auxiliary losses --- which improve the reflector and mutator without additional task executions --- beat ACE on the majority of conditions across all three benchmarks: optimizer state adds $+4.8$ TGC over ACE on AppWorld Normal/Lite and $+6.4$ accuracy on BrowseComp+/Nano; auxiliary losses add $+5.5$ over ACE on AppWorld Challenge/Lite and $+8.5$ on RewardBench2/Nano. Grouped rollouts also improves reliably, but at additional execution cost. Since the two cheapest primitives are among the most effective, diagnostic precision matters more than execution volume.

\paragraph{Execution-side primitives must be tuned to task dynamics.} Batching shows strong gains when the failure distribution is broad ($+5.4$ over ACE on AppWorld Normal/Lite) but can actively hurt when failures are diverse: on BrowseComp+/Lite, batching \textit{degrades} by $-6.0$ relative to ACE, suggesting the mutator is overloaded with competing signal. Grouped rollouts help most when the reflector needs contrastive signal: $+3.0$ over ACE on AppWorld Normal/Lite and $+15.1$ over ACE on RewardBench2/Nano (the largest single gain in the table). Credit assignment adds modest value on AppWorld, where multi-step
procedural traces benefit from entry-level attribution, but not on
BrowseComp+, where terminal feedback already localizes the problem to the
agent's search strategy rather than to individual playbook entries. These results suggest configuring execution-side primitives to the variance structure and difficulty distribution of the target environment rather than applying them uniformly.

\paragraph{The seed-to-ceiling gap shapes what kind of learning occurs.} On AppWorld, where the gap is narrow, multiple primitives contribute: batching and optimizer state lead on Normal/Lite ($+5.4$ and $+4.8$ over ACE), while auxiliary losses and optimizer state lead on Challenge/Lite ($+5.5$ and $+4.8$). Learned playbooks confirm incremental gains, accumulating targeted procedural rules via ADD mutations. On BrowseComp+, where the gap is wide, results are agent-dependent: optimizer state gives the largest gain for Nano ($+6.4$ over ACE) but slightly degrades Lite ($-1.3$), where failure replay leads instead ($+2.7$). Playbooks show a high ratio of UPDATE mutations as strategies are refined over repeated encounters. On RewardBench2, where the environment is near-deterministic,
over-instruction poses the greatest risk. Unlike AppWorld and BrowseComp+,
where the agent executes multi-step procedures, RewardBench2 is a
single-turn judgment task, and playbook entries learned from procedural
failure modes (e.g., structured trigger/procedure rules) can interfere with
the naturalistic reasoning the task requires. This style mismatch is most visible on Nano: ACE falls from 67.2 at seed to 62.7 after optimization ($-4.5$), and failure replay only partially recovers it to 65.4. Contrastive signal (grouped rollouts) provides the most effective learning signal in this regime, improving Nano from 62.7 to 77.8 ($+15.1$ over ACE, $+10.6$ over Seed), because the reflector observes both correct and incorrect rankings under the same playbook and can identify which criteria are actually predictive rather than proposing new procedural rules from failures alone.

% [RB2-TODO: add playbook characterization for RewardBench2.]

\begin{table*}[!t]
\centering
\small
\caption{\textbf{Leave-one-out ablation of the composed \rcl{} optimizer.} Each row removes one primitive from the full \rcl{} composition. Deltas are relative to full \rcl{}. Positive deltas indicate settings where removing a primitive improves over the full composition, while negative deltas indicate settings where that primitive contributes positively in composition. Largest degradation per column is \underline{underlined}.}
\label{tab:ablation-removal}
\resizebox{\textwidth}{!}{%
\begin{tabular}{@{} l cc cc cc cc @{}}
\toprule
& \multicolumn{4}{c}{\textbf{AppWorld}} & \multicolumn{2}{c}{\textbf{BrowseComp+}} & \multicolumn{2}{c}{\textbf{RewardBench2}} \\
\cmidrule(lr){2-5} \cmidrule(lr){6-7} \cmidrule(lr){8-9}
& \multicolumn{2}{c}{Normal} & \multicolumn{2}{c}{Challenge} & \multicolumn{2}{c}{Test Set} & \multicolumn{2}{c}{Test Subset} \\
\cmidrule(lr){2-3} \cmidrule(lr){4-5} \cmidrule(lr){6-7} \cmidrule(lr){8-9}
& Lite & Nano & Lite & Nano & Lite & Nano & Lite & Nano \\
\midrule
\rcl{} (all primitives) & 89.3\phantom{\up{00.0}} & 89.1\phantom{\up{00.0}} & 71.9\phantom{\up{00.0}} & 83.7\phantom{\up{00.0}} & 37.3\phantom{\up{00.0}} & 51.3\phantom{\up{00.0}} & 82.4\phantom{\up{00.0}} & 69.8\phantom{\up{00.0}} \\
\midrule
$-$ Failure Replay      & 85.7\dn{-3.6}              & 88.5\dn{-0.6}              & 71.5\dn{-0.4}              & \underline{74.9}\dn{-8.8}  & 38.7\up{1.4}               & \underline{33.3}\dn{-18.0} & 83.1\up{0.7}               & 59.4\dn{-10.4} \\
$-$ Optimizer State     & 89.9\up{0.6}               & 85.5\dn{-3.6}              & 71.2\dn{-0.7}              & 82.5\dn{-1.2}              & 34.0\dn{-3.3}              & 46.7\dn{-4.6}              & \underline{74.6}\dn{-7.8}  & 64.1\dn{-5.7} \\
$-$ Credit Assignment   & 89.9\up{0.6}               & \underline{85.0}\dn{-4.1}  & 73.4\up{1.5}               & 82.1\dn{-1.6}              & \underline{32.7}\dn{-4.6}  & 41.3\dn{-10.0}             & 74.3\dn{-8.1}              & 75.1\up{5.3} \\
$-$ Grouped Rollouts    & \underline{79.8}\dn{-9.5}  & 86.8\dn{-2.3}              & 71.9\phantom{\dn{0.0}}     & 79.2\dn{-4.5}              & 36.7\dn{-0.6}              & 45.3\dn{-6.0}              & 78.5\dn{-3.9}              & \underline{58.5}\dn{-11.3} \\
$-$ Batching            & 83.9\dn{-5.4}              & 86.3\dn{-2.8}              & \underline{70.4}\dn{-1.5}  & 79.4\dn{-4.3}              & 40.0\up{2.7}               & 36.7\dn{-14.6}             & 81.2\dn{-1.2}              & 64.6\dn{-5.2} \\
$-$ Auxiliary Losses    & 87.5\dn{-1.8}              & 89.2\up{0.1}               & 70.0\dn{-1.9}              & 82.2\dn{-1.5}              & 33.3\dn{-4.0}              & 49.3\dn{-2.0}              & 80.9\dn{-1.5}              & 82.4\up{12.6} \\
\bottomrule
\end{tabular}%
}
\end{table*}

\subsection{Primitive Interactions under Composition}
\label{sec:complementarity}

\paragraph{Standalone value does not predict compositional role.} To distinguish marginal gains from compositional role, we perform leave-one-out ablations of the full \rcl{} optimizer. Table~\ref{tab:main-results} measures the marginal value of adding a single primitive to ACE, whereas Table~\ref{tab:ablation-removal} measures that primitive's role once the full optimizer is assembled. These are not the same quantity. Auxiliary losses illustrate this clearly: as a standalone addition to ACE, they improve 7 of 8 settings, yet removing them from full \rcl{} causes only modest degradations in most conditions and even improves RewardBench2/Nano by 12.6 points. Credit assignment shows the converse pattern. As a standalone addition, it helps only 3 of 8 settings and ties one, but removing it from full \rcl{} produces the largest drop in three settings---AppWorld Normal/Nano, BrowseComp+/Lite, and RewardBench2/Lite---and causes a further 10.0 point drop on BrowseComp+/Nano. Batching shows a similar reversal on BrowseComp+/Nano: it hurts when added to ACE, but removing it from full \rcl{} causes a 14.6-point drop. Standalone gains therefore do not reliably predict a primitive's compositional role.

\paragraph{Two primitives are consistently load-bearing.} At the same time, the leave-one-out results do not imply that the composed optimizer is arbitrary. Two primitives are especially load-bearing in composition. Removing grouped rollouts hurts 7 of 8 settings and ties the remaining one; it never exceeds full \rcl{} and produces the largest drop on AppWorld Normal/Lite ($-9.5$) and RewardBench2/Nano ($-11.3$). Removing failure replay shows a similarly strong pattern, including the largest drop on AppWorld Challenge/Nano ($-8.8$) and the single largest regression in the table on BrowseComp+/Nano ($-18.0$). Batching and optimizer state also help in many settings---each removal hurts 7 of 8 conditions---but their effects are more distributed than concentrated in a single dominant regime. This is consistent with the training-dynamics analysis, where optimizer state mainly stabilizes convergence and batching mainly affects peak performance and relearning. The full optimizer therefore depends on several complementary mechanisms, even if their effects overlap enough that standalone ablations are a poor proxy for compositional contribution.

\paragraph{The remaining interactions are regime-dependent.} The remaining reversals point to regime-dependent interaction effects. BrowseComp+, which the main results characterize as a wide-gap, strategy-learning benchmark, is especially sensitive on Nano to removing failure replay ($-18.0$), batching ($-14.6$), and credit assignment ($-10.0$). The replay and batching drops fit the broader picture of a setting that requires repeated refinement of difficult failures, while the credit-assignment drop reinforces the gap between standalone and compositional value identified above. RewardBench2/Nano shows a different pattern: removing grouped rollouts sharply hurts ($-11.3$), reinforcing the importance of contrastive signal in a near-deterministic calibration task, while removing auxiliary losses ($+12.6$) and credit assignment ($+5.3$) improves performance, consistent with the broader observation that over-structuring can hurt Nano on this benchmark. AppWorld, where the seed-to-ceiling gap is narrower, shows distributed sensitivity rather than a single dominant dependency: no single primitive dominates, with the most critical removal shifting from grouped rollouts on Normal/Lite to failure replay on Challenge/Nano. Primitive contributions in context-space optimization are therefore real but non-additive. Their effect depends on the task regime, the agent, and the other mechanisms with which they are combined.

\begin{figure*}[t]
\centering
\includegraphics[width=\textwidth]{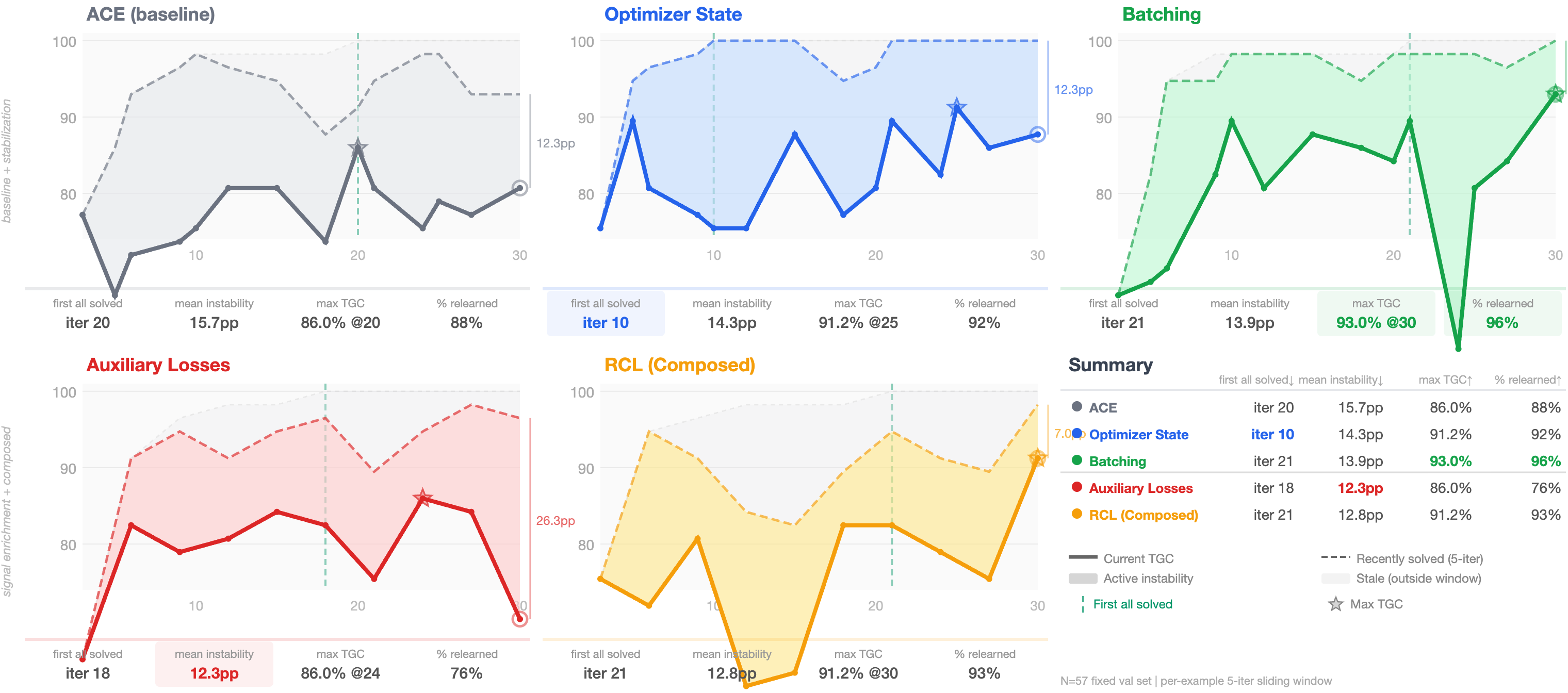}
\caption{\textbf{Learning dynamics on AppWorld dev (Gemini 3.1 Flash-Lite,
57 tasks).} Solid lines: current TGC at each checkpoint. Dashed lines:
recently solved rate (fraction of tasks solved $\geq$1$\times$ in the
trailing 5 iterations). Colored shading: \emph{active instability} ---
the gap between the recently solved rate and current TGC, measuring tasks
solved within the window but not currently retained. Gray shading:
\emph{stale regressions} --- the gap between the all-time per-example
best-so-far and the recently solved rate, measuring tasks solved
historically but not within the window. Stars: peak TGC achieved during
training. Green verticals: first iteration at which every dev task has
been solved at least once (full coverage).}%\vspace{-1em}
\label{fig:training-dynamics}
\end{figure*}

\subsection{Training Dynamics}
\label{sec:training_dynamics}

Final test-set scores (Table~\ref{tab:main-results}) show where each
primitive ends up after 30 iterations, but not how it gets there. To
understand the optimization trajectory --- whether progress is steady or
oscillatory, whether capabilities once learned are retained or forgotten
--- we track per-example solve status on a fixed 57-task AppWorld dev set
at every checkpoint for every run using the Gemini~3.1~Flash-Lite agent
(Figure~\ref{fig:training-dynamics}).

\paragraph{Metrics.} We track two quantities at each checkpoint, both
computed per-example and then aggregated. The \emph{current TGC} (solid
lines) is the fraction of dev tasks solved at that checkpoint. The
\emph{recently solved rate} (dashed lines) is the fraction of dev tasks
solved at least once in a trailing window of $w{=}5$ iterations: for each
task, we check whether any checkpoint in the window solves it, then take
the fraction across tasks. The recently solved rate is always $\geq$ the
current TGC, because any task solved now was also solved within the
window. This rate can decrease over time as tasks fall out of the trailing
window.

The gap between the recently solved rate and the current TGC decomposes
into two components, visualized as shading in
Figure~\ref{fig:training-dynamics}. \emph{Active instability} (colored
shading) is this gap directly: the fraction of tasks the optimizer has
solved within the last $w$ iterations but that the current playbook fails
on. These are recent regressions --- capabilities demonstrated within the
window but not currently retained. \emph{Stale regressions} (gray
shading) capture a second layer: the gap between the all-time per-example
best-so-far envelope and the recently solved rate. These are tasks solved
at some point historically but not within the trailing window ---
capabilities lost earlier in training and not yet recovered. Together, the
two shadings decompose the full gap between demonstrated capability
(all-time envelope) and current performance into recent forgetting
(active) and older forgetting (stale).

We report four summary statistics per run.
\emph{First all solved}: the first iteration at which every dev task has
been solved at least once (cumulative, all-time). This does not mean all
tasks are solved simultaneously --- it means the optimizer has, at some
point, produced a playbook capable of solving each individual task.
\emph{Mean instability}: the mean active instability gap across all
checkpoints.
\emph{Max TGC}: the highest current TGC at any single checkpoint.
\emph{\% relearned}: when a task flips from solved to unsolved between
consecutive checkpoints, that is an unlearn event; if it later flips back
to solved, that is a recovery. \% relearned is the fraction of unlearn
events that are eventually recovered, measuring how often forgetting is
reversible.

\paragraph{Results.} All primitives eventually achieve full coverage ---
every dev task is solved at least once --- but they differ sharply in
\textit{when} they reach it and \textit{how much} they retain afterward.
Optimizer State reaches full coverage earliest (iteration~10) and
achieves high peak TGC (91.2\%) with strong relearning (92\%): the rolling
state document prevents the mutator from reverting useful edits, providing
a stabilizing effect analogous to momentum in parameter-space
optimization. Batching reaches full coverage later (iteration~21) and
exhibits larger mid-training oscillations, but achieves the highest peak
TGC of any configuration (93.0\%) with the highest relearning rate
(96\%) --- consistent with the variance-reduction interpretation, where
larger batches produce noisier signal early but increasingly robust signal
as the failure distribution becomes well-characterized.
Auxiliary Losses presents a distinctive profile: it achieves the
\textit{lowest} mean instability (12.3pp) but also the lowest relearning
rate (76\%) and a peak TGC (86.0\%) no higher than the ACE baseline. This
suggests a conservative rather than exploratory dynamic --- structured
diagnostics produce stable, targeted edits that rarely regress, but when
regressions do occur they tend to be permanent, and the primitive discovers
fewer novel solutions overall.
The composed \rcl{} configuration inherits complementary strengths: low
instability (12.8pp, second only to Auxiliary Losses), high peak TGC
(91.2\%, matching Optimizer State), and strong relearning (93\%),
consistent with the complementary-pathology view from Section~\ref{sec:main_results}.

% =============================================
% REWRITTEN Section 4.4 + Figure 3 caption
% =============================================

% ---- FIGURE 3 CAPTION (replaces the current one) ----

% ---- SECTION 4.4 (replaces current text) ----

% =============================================
% Section 4.4 — arxiv preprint version (expanded)
% =============================================

% =============================================
% Section 4.4 — arxiv preprint version (expanded)
% =============================================

% =============================================
% Subsections within Section 4 (Experiments) — arxiv preprint
% These follow Training Dynamics (4.3) as 4.4, 4.5, 4.6
% =============================================

% =============================================
% Revised Sections 4.4–4.6
% =============================================
 
\subsection{Sensitivity to Initialization}
\label{sec:seed}
 
\begin{figure*}[t]
\centering
\includegraphics[width=\textwidth]{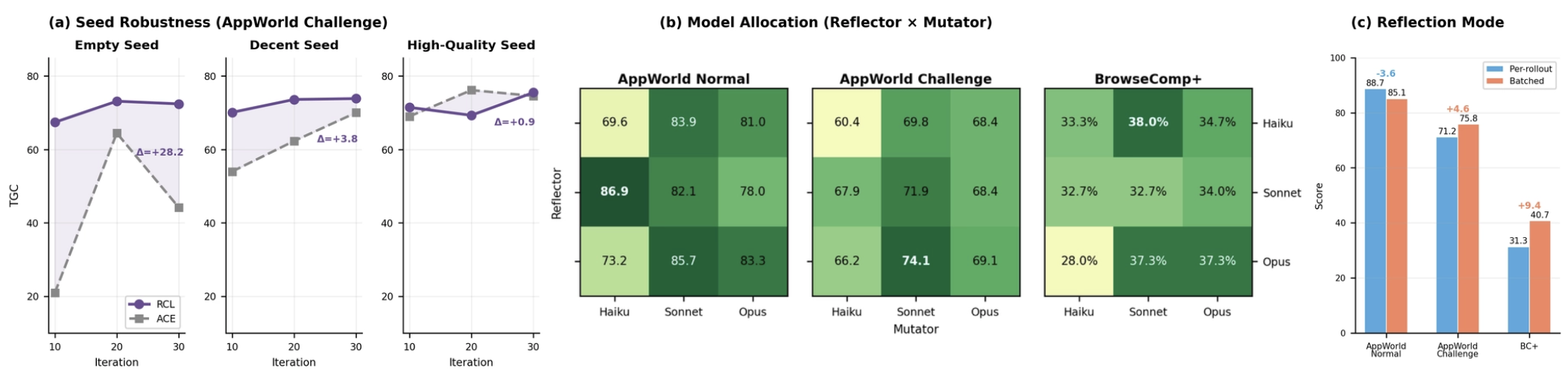}
\caption{\textbf{Design choice analysis (Gemini 3.1 Flash-Lite).}
\textbf{(a)}~Seed robustness on AppWorld Challenge: RCL converges to 72--76 TGC
from all three seeds; ACE without primitives diverges from weaker initializations.
\textbf{(b)}~Reflector $\times$ mutator allocation across benchmarks.
Performance depends on the interaction between both roles and the task regime;
no single configuration dominates uniformly.
\textbf{(c)}~Per-trace vs.\ batched reflection with $B{=}3$.
Batched reflection helps on harder tasks (AW Challenge, BC+) but
hurts when failures are diverse.}
\vspace{-1em}
\label{fig:multi-plot}
\end{figure*}

Figure~\ref{fig:multi-plot}a varies the seed playbook across three quality
levels: (I) empty (no entries), (II) decent (7 entries across 4 sections), and (III) high-quality (9 entries across 5 sections) on AppWorld Challenge. For more details please see Appendix~\ref{app:quality-playbooks}. RCL converges to 72--76 TGC from all three,
while ACE without primitives oscillates severely from the empty seed (44.2
at iteration~30 vs.\ RCL's 72.4).
 
\textbf{Primitive contribution scales inversely with seed quality:}
$+28.2$ from the empty seed, $+3.8$ from the decent seed, and
$+0.9$ from the high-quality seed. This pattern is consistent with the
primitives addressing genuine optimization pathologies --- variance,
forgetting, and instability --- whose effects are largest when
initialization is weak. From an empty seed, ACE must both discover useful
rules and retain them across iterations, so noisy updates and regressions
are especially costly; the added primitives make those early improvements
easier to accumulate. From a strong seed, the remaining errors are fewer
and more local, so the marginal value of these stabilizing mechanisms
shrinks. ACE's divergence from the empty initialization is consistent with
this interpretation.
 
\subsection{Model Allocation}
\label{sec:model-allocation}
 
Figure~\ref{fig:multi-plot}b varies the reflector and mutator models independently across all nine combinations of Haiku, Sonnet, and Opus. Stronger reflectors tend to help on harder tasks: Opus reflector with Sonnet mutator achieves 74.1 on AppWorld Challenge, the best configuration on that split. However, the pattern is not monotonic in overall model capability. On AppWorld Normal and BrowseComp+, Haiku as reflector paired with Sonnet as mutator matches or exceeds several Opus-reflector configurations. More strikingly, Opus as mutator does not dominate despite being the strongest model; Sonnet is the most consistently strong mutator across benchmarks.
We hypothesize this reflects a difference in what the two roles demand: reflection requires diagnostic reasoning over multi-step failures, while mutation requires faithful, constrained editing — and a model that is too capable in the latter role may over-interpret the diagnostic rather than execute it precisely. The consistency of this pattern across three benchmarks with different task structures suggests that matching the reflector's output complexity to the mutator's execution capacity matters more than uniformly maximizing capability.
 
\subsection{Per-Trace vs.\ Batched Reflection}
\label{sec:reflection-aggregation}

In Section~\ref{sec:batching}, batching was defined as an execution-side
primitive: sampling $B$ tasks per iteration and presenting their
diagnostics to the mutator. A separate design choice is \textit{where}
aggregation occurs --- whether the reflector sees all $B$ traces in a
single call (batched reflection) or reflects on each independently, with
aggregation deferred to the mutator. Figure~\ref{fig:multi-plot}c compares
these modes at $B{=}3$. Batched reflection improves over per-trace
reflection on harder tasks (AppWorld Challenge $+4.6$, BrowseComp+ $+9.4$),
but degrades on AppWorld Normal ($-3.6$).

A useful way to interpret per-trace reflection is as producing a noisy
directional update for each trace, analogous to a stochastic gradient
estimate. The mutator then performs the context-space analogue of a
minibatch step by reconciling multiple $\Delta_i$ into a single playbook
update. Batched reflection instead asks the reflector to estimate a shared
update direction directly from several traces at once. This can help when
failures are coherent enough to support cross-trace synthesis, as on
harder tasks, but it can hurt when the remaining failures are diverse and
require more localized corrections.

This distinction also clarifies the interaction with the batching result
from Table~\ref{tab:main-results}, where batching degrades BrowseComp+ by
$-6.0$ relative to ACE. The key difference is \textit{where} reconciliation
occurs: in batching, multiple independent diagnostics are passed to the
mutator, which must reconcile them; in batched reflection, the reflector
synthesizes across traces and passes a single coherent signal. Aggregation
at the reflector helps on BrowseComp+ (relative to per-trace) while
aggregation at the mutator hurts (relative to ACE), suggesting that in
regimes with diverse failures, the mutator's capacity to reconcile
competing recommendations --- not the volume of signal --- is the binding
constraint.

For the main experiments, we therefore use per-trace reflection not
because batched reflection is uniformly worse, but because it preserves the
modular RCL decomposition and composes more cleanly with grouped rollouts,
whose contrastive signal is defined at the individual-task level.

% \paragraph{Composition is not additive.}
% The full \rcl{} configuration does not uniformly dominate individual
% primitives in Table~\ref{tab:main-results}, suggesting that some primitives
% interact destructively when composed. Diagnosing these interactions
% systematically is an important open direction.

\section{Conclusion}
\label{sec:conclusion}

\looseness=-1
As agents take on increasingly complex tasks and operate in increasingly diverse environments, the ability to learn from experience through context updates --- rather than weight updates --- will become a core capability. A growing body of work has begun developing methods for this setting, but these methods have been studied in isolation, each under different models, benchmarks, and prompting conventions, making it difficult to attribute improvements to specific learning mechanisms. Reflective Context Learning (\rcl{}) addresses this by treating context-space adaptation as a single optimization problem: the same pathologies that arise in parameter-space learning (variance, forgetting, local optima) arise in context space, and the same class of remedies apply. By recasting recent context-optimization methods as instances of a shared loop and studying their primitives under controlled conditions, we identify four findings.
\textbf{(1)}~Diagnostic precision matters more than execution volume: primitives that improve the reflection signal give the largest gains per unit of compute.
\textbf{(2)}~Which primitives help depends on the task regime, and composition is not additive: no single primitive dominates, and the full optimizer does not uniformly beat the best individual one.
\textbf{(3)}~Matching model capacity to each role matters more than maximizing it: a faithful mutator paired with a strong reflector outperforms the reverse.
\textbf{(4)}~Context-space training dynamics mirror parameter-space phenomena: oscillation, momentum-stabilized convergence, and a tradeoff between stability and relearning.

\looseness=-1
More broadly, these findings suggest that context-space optimization will benefit from the same systematic discipline that classical ML brings to weight updates: diagnosing pathologies, composing remedies, and studying their interactions. Several directions remain open. Adaptive primitive selection --- choosing which primitives to activate based on the current training phase or task properties --- could reduce the need for manual configuration. Second-order state tracking, where the optimizer reasons about the trajectory of its own edits rather than just the current batch, may further stabilize convergence. Extension to continual deployment, where the task distribution shifts over time and the playbook must adapt without forgetting, is a natural next step. As models grow more capable, the scope of what can be learned through context updates grows with them --- making principled optimization of that learning process increasingly important.
\bibliography{colm2026_conference}
\bibliographystyle{colm2026_conference}

\clearpage
\appendix
%============================================================
%APPENDIX MATERIAL
%===========================================================
\section{Design Rationale and Future-Proofing}
\label{app:design_rationale}
Several choices in the RCL formulation of Section~\ref{sec:rcl:formulation} are deliberate and designed to accommodate increasingly capable models.
\paragraph{Independent components.} The reflector $g$, mutator $f$, and agent $\mathcal{A}$ are defined as independent components that may be instantiated by the same model or by different models of varying capability. This separation reflects a practical reality: the cognitive demands of diagnosing why a strategy failed (reflection) are qualitatively different from the demands of executing a constrained text edit (mutation) or executing tasks in an environment (forward pass). Allocating model capacity independently to each role allows the system to improve as any individual component improves.
\paragraph{Structured context artifact.} We formulate $\mathcal{C}$ as a structured, modular artifact rather than a flat text string. A structured playbook with individually addressable rules enables \textit{localized} updates: the reflector can identify that a specific rule caused a failure and propose a targeted revision, while a flat prompt can only be rewritten holistically. The granularity of the learned artifact determines the granularity of credit assignment that is possible, just as the choice of network architecture in classical learning determines the granularity of gradient updates.
\paragraph{Expanding scope with model capability.} These choices are designed to be future-proof. As base models improve, the reflector can produce more precise diagnoses, the mutator can execute more sophisticated edits, and the scope of what $\mathcal{C}$ can encode expands from phrasing adjustments to genuine multi-step strategies, tool definitions, and structured behavioral policies. The formulation imposes no ceiling on what can be learned through context updates; it only requires that the learning signal be available to evaluate trajectory quality.
%============================================================
%APPENDIX TABLE: Per-method RCL mapping
%===========================================================
\begin{table*}[t]
\centering
\footnotesize
\caption{Prior context-optimization methods as instances of the RCL loop, grouped by developmental thread.}
\label{tab:prior-methods-full}
\begin{tabular}{@{} p{3.5cm} p{2.8cm} p{2.8cm} p{2.6cm} p{1.8cm} @{}}
\toprule
\textbf{Method} & \textbf{Reflector ($g$)} & \textbf{Mutator ($f$)} & \textbf{State / Memory} & \textbf{Regime} \\
\midrule
\multicolumn{5}{@{}l}{\textit{Loop development}} \\[2pt]
Reflexion \citep{shinn2023reflexion} & Verbal self-critique & Append to memory & Episodic history & Single-ep. \\
ExpeL \citep{zhao2024expel} & Experience extraction & Insight reuse & Extracted knowledge & Cross-task \\
Agent-Pro \citep{zhang2024agentpro} & Belief + policy critique & DFS-style search & World model beliefs & Policy \\
Dyn.\ Cheatsheet \citep{suzgun2026dynamic} & Failure summarization & Curated append & Persistent cheatsheet & Task learn. \\
ACE \citep{zhang2026ace} & Trajectory critique & Structured delta edits & Playbook history & Policy \\
\midrule
\multicolumn{5}{@{}l}{\textit{Reflection mechanism}} \\[2pt]
ProTeGi \citep{pryzant2023automatic} & Minibatch text gradient & Beam search + bandit & None & Micro-opt. \\
TextGrad \citep{yuksekgonul2024textgrad} & Textual differentiation & LLM-proposed edit & None & Modular opt. \\
ERM \citep{yan2025efficient} & Exemplar reflection & Beam search & Historical feedback & Micro-opt. \\
TF-GRPO \citep{cai2025trainingfreegrpo} & Semantic group advantage & Context update & Experiential library & Task learn. \\
\citet{ding2025samplingmomentum} & Sampling momentum & Textual grad.\ descent & Past distributions & Micro-opt. \\
\midrule
\multicolumn{5}{@{}l}{\textit{Search and selection}} \\[2pt]
APE \citep{zhou2023ape} & None (score only) & Monte Carlo selection & None & Instr.\ search \\
EvoPrompt \citep{guo2024evoprompt} & None (score only) & Evolutionary mutation & Candidate pop. & Instr.\ search \\
PromptBreeder \citep{fernando2024promptbreeder} & None (score only) & Self-referential & Meta-prompts & Instr.\ search \\
GEPA \citep{agrawal2025gepa} & Pareto-aware reflection & Evolutionary + Pareto & Candidate pop. & Policy \\
\midrule
\multicolumn{5}{@{}l}{\textit{Structured program optimization}} \\[2pt]
DSPy \citep{khattab2024dspy} & N/A (compiler) & Module-level compilation & Program structure & Program opt. \\
MIPRO \citep{opsahl-ong-etal-2024-optimizing} & Proposal scoring & Bayesian surrogate & Module-level state & Program opt. \\
\bottomrule
\end{tabular}
\end{table*}

% \section{Primitive Taxonomy}
% \label{app:primitive-taxonomy}

% Table~\ref{tab:primitives-app} summarizes the six optimization primitives studied in this work, the classical pathology each addresses, the stage of the RCL loop it targets, and partial prior instantiations in context space.

% \begin{table}[h]
% \centering
% \small
% \caption{Optimization primitives, the classical pathologies they address, and the RCL loop stage each targets.}
% \label{tab:primitives-app}
% \begin{tabular}{@{} p{3.1cm} p{1.4cm} p{4.8cm} p{3.3cm} @{}}
% \toprule
% \textbf{Primitive} & \textbf{Stage} & \textbf{Pathology} & \textbf{Prior Work} \\
% \midrule
% Batching & Execution & High variance from single samples & ProTeGi; TF-GRPO \\
% \lightrule
% Grouped Rollouts & Execution & Confounded attribution & TF-GRPO \\
% \lightrule
% Credit Assignment & Reflection & Sparse terminal reward & TextGrad \\
% \lightrule
% Auxiliary Losses & Reflection & Surface-level diagnostics & --- \\
% \lightrule
% Failure Replay & Sampling & Forgetting learned tactics & Dyn.\ Cheatsheet; ExpeL \\
% \lightrule
% Optimizer State & Mutation & Oscillation from stateless updates & ERM; OPRO \\
% \bottomrule
% \end{tabular}
% \end{table}

\section{Extended Training Dynamics}
\label{app:training-dynamics}

Section~\ref{sec:training_dynamics} defines two forms of instability ---
active (recent regressions within the trailing window) and stale (older
regressions outside the window) --- using a window of $w{=}5$ iterations.
The window size controls how strictly we define ``recent'': a shorter
window requires a task to have been solved very recently to avoid being
counted as stale, while a longer window is more forgiving. Varying the
window reveals whether a primitive maintains capabilities through
sustained re-solving (narrow active gap even at short windows) versus
relying on one-off discoveries that are not reproduced (active gap widens
as the window shrinks, with more instability shifting from active to
stale). We show this analysis for all primitives under three window sizes.

\begin{figure*}[t]
\centering
\includegraphics[width=\textwidth]{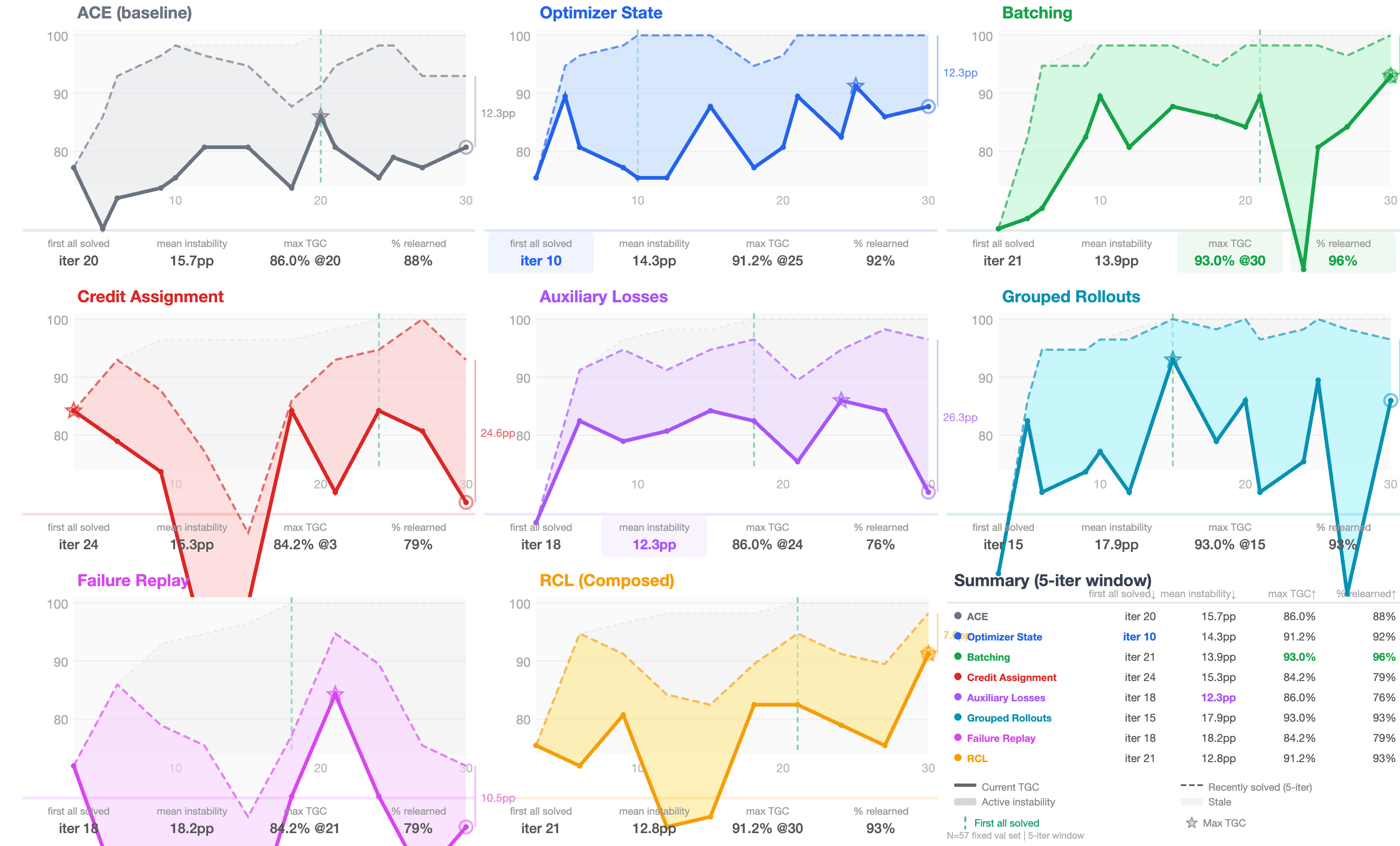}
\caption{Training dynamics for all primitives with a 5-iteration sliding
window. Format follows Figure~\ref{fig:training-dynamics}: solid = current
TGC; dashed = recently solved rate (per-example union over 5 iterations);
colored shading = active instability (solved within window but not now);
gray shading = stale regressions (solved historically but not within
window). This is the strictest recency condition and produces the widest
active instability gaps.}
\label{fig:dynamics-w5}
\end{figure*}

\paragraph{5-iteration window (Figure~\ref{fig:dynamics-w5}).} This is the
strictest recency condition and produces the widest active instability
gaps: a task must have been solved within the last 5 checkpoints to avoid
being classified as stale. Under this view, Optimizer State and Grouped
Rollouts show the narrowest active gaps, indicating they retain
capabilities through consistent re-solving. Failure Replay shows the
widest active gaps but also the highest \% relearned --- tasks that
regress are quickly re-encountered via the replay buffer and re-solved,
producing high volatility but reliable recovery.

\begin{figure*}[t]
\centering
\includegraphics[width=\textwidth]{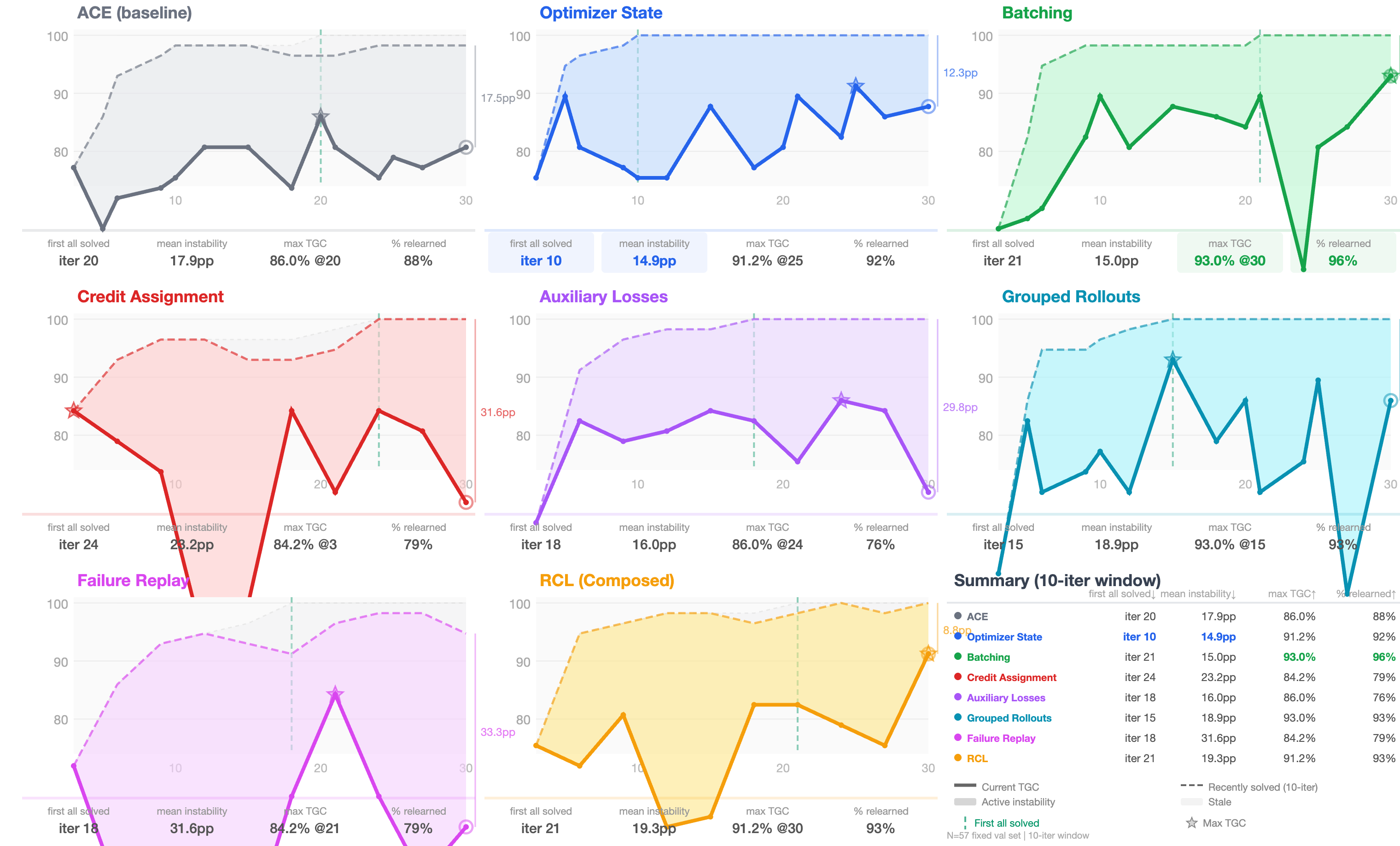}
\caption{Training dynamics with a 10-iteration sliding window. Tasks
solved anywhere in the last 10 iterations count as recently solvable,
shifting instability from the active to the stale category. Active gaps
narrow relative to the 5-iteration view (Figure~\ref{fig:dynamics-w5}),
but the relative ordering of primitives is preserved.}
\label{fig:dynamics-w10}
\end{figure*}

\paragraph{10-iteration window (Figure~\ref{fig:dynamics-w10}).} The
wider window is more forgiving: tasks solved anywhere in the last 10
iterations count as recently solvable, shifting instability from the
active to the stale category. Active gaps narrow across all primitives
relative to the 5-iteration view, but the relative ordering is preserved.
This confirms that the differences between primitives reflect genuine
differences in retention behavior, not artifacts of the window size.

\begin{figure*}[t]
\centering
\includegraphics[width=\textwidth]{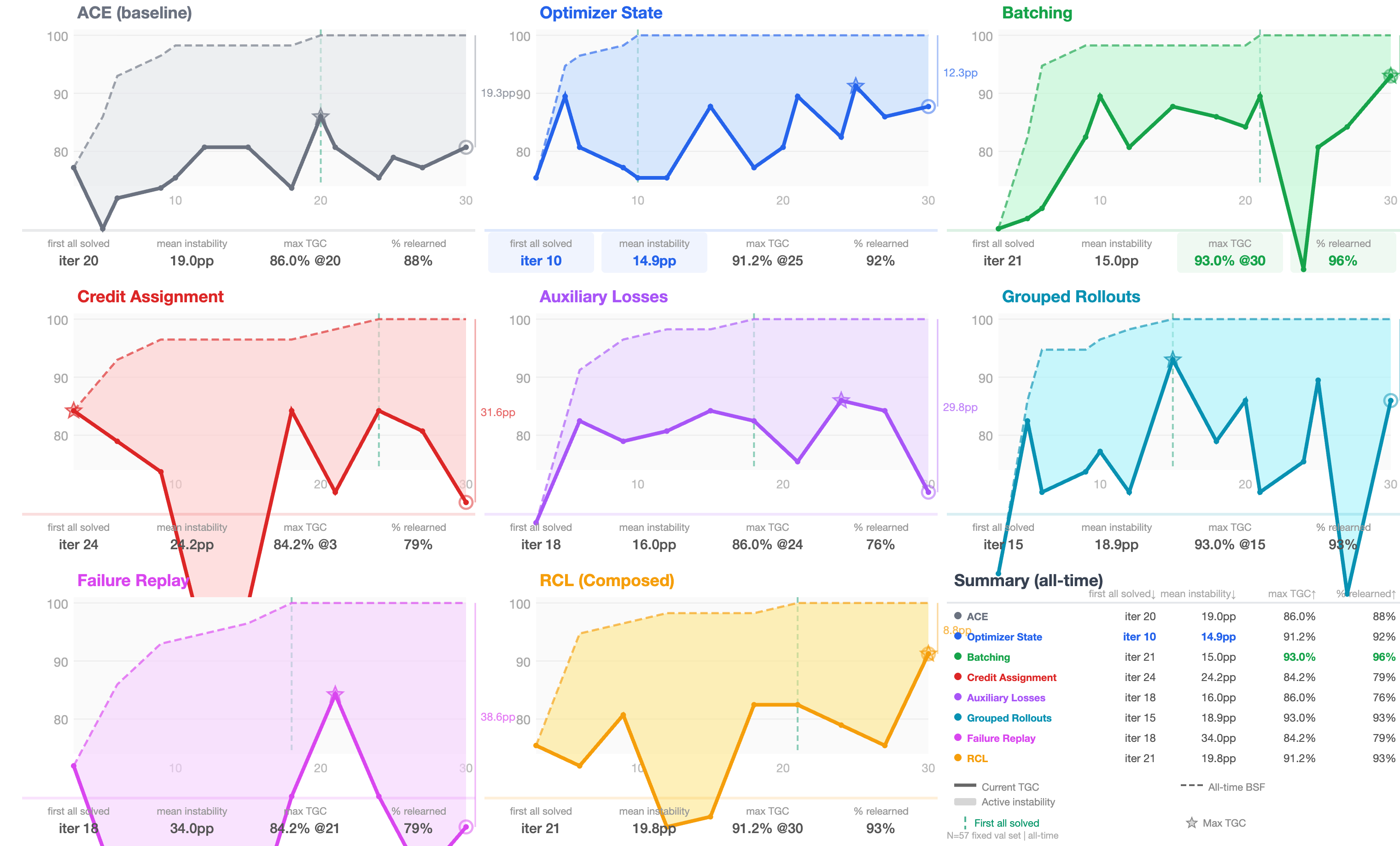}
\caption{Training dynamics with the all-time best-so-far envelope (no
window). The dashed line is now monotonically non-decreasing and converges
to 100\% for all runs. All instability is active by definition (the stale
category is empty). The remaining gap measures capabilities ever
demonstrated but not currently retained. This is the most permissive view
and produces the narrowest gaps.}
\label{fig:dynamics-alltime}
\end{figure*}

\paragraph{All-time envelope (Figure~\ref{fig:dynamics-alltime}).} With no
window, the recently solved rate becomes the cumulative best-so-far: a
task counts as recently solvable if it was ever solved. The dashed line is
now monotonically non-decreasing and converges to 100\% for all runs,
confirming that every primitive eventually discovers a playbook capable of
solving each dev task. All instability is now active by definition (the
stale category is empty), and the remaining gap --- between the all-time
envelope and current TGC --- is the classical forgetting measure:
capabilities ever demonstrated but not currently retained. Even under this
most permissive view, meaningful gaps persist for all primitives except
Optimizer State and Grouped Rollouts, underscoring that catastrophic
forgetting is a genuine pathology in context-space optimization.

\section{Prompts}
\label{app:prompts}
This section provides the prompts used for AppWorld (Figure~\ref{fig:appworld-system-prompt}), BrowseComp+ (Figure~\ref{fig:browsecomp-system-prompt}), and RewardBench2 (Figure~\ref{fig:rewardbench2-system-prompt}).

\begin{figure}[h]
\begin{promptbox}{AppWorld System Prompt}
I am your supervisor, and you are an AI Assistant whose job is to complete my day-to-day tasks fully autonomously.

To do this, you will need to interact with app(s) (e.g., spotify, venmo etc) using their associated APIs on my behalf. Use the \texttt{execute\_python} tool to run code one step at a time, observe the output, then decide the next step.

Key APIs:

\texttt{execute\_python(code="print(apis.api\_docs.show\_app\_descriptions())")}

\texttt{execute\_python(code="print(apis.api\_docs.show\_api\_descriptions(app\_name='spotify'))")}

\texttt{execute\_python(code="print(apis.api\_docs.show\_api\_doc(app\_name='spotify', api\_name='login'))")}

\textbf{Key instructions}:

A. General instructions:

- Act fully on your own. Never ask for confirmation.

- Never invent or guess values - always look them up via APIs.

- Never leave placeholders - always fill in real values.

B. App-specific instructions:

- All credentials are in the Supervisor app: \texttt{apis.supervisor.show\_account\_passwords()}

- References to friends/family refer to phone contacts.

- Get current date/time from \texttt{datetime.now()} or phone API, never your internal clock.

- Paginated APIs: Always loop through all pages. Don't stop at first page.

C. Code-operation instructions:

- Use the \texttt{execute\_python} tool to run code. Execute one step at a time, observe its output, then decide the next step.

- Always use \texttt{print()} to see API return values - otherwise output is hidden.

- Always check API specs before calling an API.

D. Task-completion:

- Call \texttt{apis.supervisor.complete\_task(answer=...)} when done.

- Keep answers minimal (just the value, not full sentences).

- Numbers must be numeric ("10" not "ten").

\# Playbook

You have been given a curated playbook of strategies, common mistakes, and proven solutions. Read it carefully and actively apply its guidance throughout your task execution.

\textit{[Dynamically generated playbook entries are appended here during evaluation.]}
\end{promptbox}
\caption{Base system prompt for AppWorld benchmark. The playbook section header is followed by dynamically generated playbook entries during evaluation.}
\label{fig:appworld-system-prompt}
\end{figure}

\begin{figure}[h]
\begin{promptbox}{BrowseComp System Prompt}
You are a deep research agent. You need to answer the given question by interacting with a search engine, using the search tool provided. Please perform reasoning and use the tool step by step, in an interleaved manner. You may use the search tool multiple times.

Your response should be in the following format:

Explanation: \{\{your explanation for your final answer. Cite evidence documents inline by enclosing their docids in square brackets [].\}\}

Exact Answer: \{\{your succinct, final answer\}\}

Confidence: \{\{your confidence score between 0\% and 100\% for your answer\}\}

\# Playbook

You have been given a curated playbook of strategies, common mistakes, and proven solutions. Read it carefully and actively apply its guidance throughout your search and reasoning.

\textit{[Dynamically generated playbook entries are appended here during evaluation.]}
\end{promptbox}
\caption{Base system prompt for BrowseComp benchmark. The playbook section header is followed by dynamically generated playbook entries during evaluation.}
\label{fig:browsecomp-system-prompt}
\end{figure}

\begin{figure}[h]
\begin{promptbox}{RewardBench2 System Prompt}
You are evaluating candidate assistant responses for the same user prompt.

Your job is to act like a careful reward model judge:

- score each candidate independently rather than rewarding position or style alone

- prefer responses that are correct, instruction-following, relevant, complete, and appropriately safe

- penalize hallucinations, mathematical mistakes, broken constraints, evasive padding, unsafe compliance, and bad refusals

- only treat responses as tied if they are genuinely equally strong overall

Return ONLY valid JSON with this schema:

\texttt{\{}

\texttt{~~"reasoning": "2-5 concise sentences explaining the main quality differences",}

\texttt{~~"ratings": [0.0, 0.0],}

\texttt{~~"best\_response\_ids": ["1"]}

\texttt{\}}

Rules:

- ratings must be numeric scores from 0.0 to 10.0

- ratings must be a JSON array containing exactly one score per candidate, in candidate-id order

- "best\_response\_ids" must contain the id(s) with the highest score

- do not include markdown fences or any text outside the JSON object

\# Playbook

You have been given a curated playbook of judging strategies, failure patterns, and calibration rules. Read it carefully and apply it while rating candidates.

\textit{[Dynamically generated playbook entries are appended here during evaluation.]}
\end{promptbox}
\caption{Base system prompt for RewardBench2 benchmark. The playbook section header is followed by dynamically generated playbook entries during evaluation.}
\label{fig:rewardbench2-system-prompt}
\end{figure}

\section{Seed Playbooks}
\newpage

\label{app:seed-playbooks}
This section provides the seed playbooks for which we initialized the ACE and RCL experiments in AppWorld (Figure ~\ref{fig:playbook-seed-appworld}),and RewardBench2 (Figure ~\ref{fig:playbook-seed-rb2}). For Browsecomp+, we initialize the playbook with no entries.
\begin{figure}[htbp]
\begin{promptbox}{Seed Playbook -- AppWorld}
\textbf{strategies\_and\_hard\_rules} (3 entries)\newline
Always end code blocks with \texttt{```} on its own line. Never use ellipsis (...) in code.\newline
Write small, incremental code chunks -- verify each step works before proceeding.\newline
Variables persist across code blocks -- reuse them instead of re-fetching.
\newpage
\textbf{apis\_to\_use\_for\_specific\_information} (3 entries)\newline
Check API docs with \texttt{apis.api\_docs.show\_api\_doc(app\_name, api\_name)} before calling any API.\newline
Use \texttt{supervisor} app for account credentials: \texttt{apis.supervisor.show\_account\_passwords()}\newline
Use \texttt{phone} app for contact/person lookup: \texttt{apis.phone.show\_contacts()} to find friends/family.

\textbf{useful\_code\_snippets\_and\_templates} (1 entries)\newline
Paginated APIs require looping: \texttt{while True: page = api(page\_index=i); if not page: break}

\textbf{common\_mistakes\_and\_correct\_strategies} (1 entries)\newline
Example credentials shown in few-shot examples are FAKE -- always fetch real credentials from supervisor.

\textbf{verification\_checklist} (1 entries)\newline
Always call \texttt{apis.supervisor.complete\_task()} when done -- this marks the task complete. Only pass an \texttt{answer=} argument if the task explicitly asks you to find, calculate, or return a specific value. For action-only tasks (e.g., \texttt{send a message}, \texttt{play a song}), call \texttt{complete\_task()} with NO answer argument.
\end{promptbox}
\caption{Seed Playbook -- AppWorld. Contains 9 entries across 5 sections.}
\label{fig:playbook-seed-appworld}
\end{figure}

\begin{figure}[htbp]
\begin{promptbox}{Seed Playbook -- RewardBench2}
\textbf{core\_ranking\_criteria} (1 entries)\newline
Score each candidate independently against the user prompt before comparing candidates to each other.

\textbf{instruction\_following\_checks} (1 entries)\newline
When the prompt contains explicit constraints, verify each candidate against every constraint instead of rewarding vague relevance.

\textbf{factuality\_and\_math\_checks} (1 entries)\newline
Penalize factual or mathematical mistakes even if the answer sounds polished or confident.

\textbf{safety\_and\_refusal\_calibration} (1 entries)\newline
Reward appropriate refusals only when the request is actually unsafe or disallowed; otherwise penalize unnecessary refusal.

\textbf{tie\_handling\_and\_score\_calibration} (1 entries)\newline
For multiple valid answers, keep the scores of correct answers close together and reserve larger gaps for separating correct from incorrect responses.

\textbf{common\_mistakes} (1 entries)\newline
Do not let verbosity, politeness, or formatting polish outweigh correctness, constraint satisfaction, and relevance.
\end{promptbox}
\caption{Seed Playbook -- RewardBench2. Contains 6 entries across 6 sections.}
\label{fig:playbook-seed-rb2}
\end{figure}

\section{Quality Playbooks on AppWorld}
For these experiments, we define the different qualities of playbook initialization for learning the dynamics of AppWorld task. We experiment with an empty playbook with 0 entries, decent playbook (Figure~\ref{fig:playbook-seed-appworld-decent}), and high quality (Figure~\ref{fig:playbook-seed-appworld-high-quality}).
\label{app:quality-playbooks}
\begin{figure}[htbp]
\begin{promptbox}{Seed Playbook (Decent) -- AppWorld}
\textbf{apis\_to\_use\_for\_specific\_information} (3 entries)

Check API docs with \texttt{apis.api\_docs.show\_api\_doc(app\_name, api\_name)} before calling any API.

Use \texttt{supervisor} app for account credentials: \texttt{apis.supervisor.show\_account\_passwords()}

Use \texttt{phone} app for contact/person lookup: \texttt{apis.phone.show\_contacts()} to find friends/family.

\textbf{strategies\_and\_hard\_rules} (2 entries)

Write small, incremental code chunks - verify each step works before proceeding.

Variables persist across code blocks - reuse them instead of re-fetching.

\textbf{useful\_code\_snippets\_and\_templates} (1 entries)

Paginated APIs require looping: \texttt{while True: page = api(page\_index=i); if not page: break}

\textbf{verification\_checklist} (1 entries)

Always call \texttt{apis.supervisor.complete\_task()} when done - this marks the task complete. Only pass an \texttt{answer=} argument if the task explicitly asks you to find, calculate, or return a specific value. For action-only tasks (e.g., `send a message`, `play a song`), call \texttt{complete\_task()} with NO answer argument.

\end{promptbox}
\caption{Seed Playbook (Decent) -- AppWorld. Contains 7 entries across 4 sections.}
\label{fig:playbook-seed-appworld-decent}
\end{figure}

\begin{figure}[htbp]
\begin{promptbox}{Seed Playbook (High Quality) -- AppWorld}
\textbf{strategies\_and\_hard\_rules} (3 entries)

Always end code blocks with \texttt{```} on its own line. Never use ellipsis (...) in code.

Write small, incremental code chunks - verify each step works before proceeding.

Variables persist across code blocks - reuse them instead of re-fetching.

\textbf{apis\_to\_use\_for\_specific\_information} (3 entries)

Check API docs with \texttt{apis.api\_docs.show\_api\_doc(app\_name, api\_name)} before calling any API.

Use \texttt{supervisor} app for account credentials: \texttt{apis.supervisor.show\_account\_passwords()}

Use \texttt{phone} app for contact/person lookup: \texttt{apis.phone.show\_contacts()} to find friends/family.

\textbf{useful\_code\_snippets\_and\_templates} (1 entries)

Paginated APIs require looping: \texttt{while True: page = api(page\_index=i); if not page: break}

\textbf{common\_mistakes\_and\_correct\_strategies} (1 entries)

Example credentials shown in few-shot examples are FAKE - always fetch real credentials from supervisor.

\textbf{verification\_checklist} (1 entries)

Always call \texttt{apis.supervisor.complete\_task()} when done - this marks the task complete. Only pass an \texttt{answer=} argument if the task explicitly asks you to find, calculate, or return a specific value. For action-only tasks (e.g., `send a message`, `play a song`), call \texttt{complete\_task()} with NO answer argument.

\end{promptbox}
\caption{Seed Playbook (High Quality) -- AppWorld. Contains 9 entries across 5 sections.}
\label{fig:playbook-seed-appworld-high-quality}
\end{figure}

\end{document}